\newcommand\SEKIusersusepackages{\usepackage{latexsym,amssymb}}
\newcommand\SEKIREVIEWSTATUS{1}
\newcommand\SEKIREVIEWER
{\brownchadname
\\FR Informatik, Universit\"at des Saarlandes, D--66123 Saarbr\"ucken, Germany
\\E-mail: {\tt cebrown@ags.uni-sb.de}
\\WWW: \url{http://www.ags.uni-sb.de/~cebrown/}
}
\newcommand\SEKIDOCUMENTCLASS{0}
\newcommand\SEKIYEAR{2005}
\newcommand\SEKINUMBEROFYEAR{01}
\newcommand\SEKITYPE{0}
\newcommand\SEKIPUBLISHEDTYPE{0}
\title{\math{\lim\tight+}, \deltaplus, 
and \NonPermutability\ of \math\beta-Steps}

\author{\wirthname
\\\Institute\\\tt\email}

\newcommand\daspaper{paper}

\newcommand\whatproofisabout{the sum of limits is the limit of the sum}
\newcommand\strongexpansionrule[6]{\LINEmath{\begin
{array}[t]{@{}c@{}l@{\mbox{~~~~~~}}l@{}}#1&&#3\\\cline
{1-1}\mediumheadroom#2&&#4\\\end{array}}}
\newcommand\branchingstrongexpansionrule[7]
{\hfill\math{
#1\over{\rule{0ex}{1.2ex}\mbox{}\hfill#2\atop\rule{0ex}{1.2ex}\mbox{}\hfill#7}}}

\newcommand\shortstrongexpansionrule[6]%
{\hfill\math{#1\over\rule{0ex}{1.7ex}#2}}

\newenvironment{slide}{\yestop\par\noindent\ignorespaces}{\par}

\newcommand\kernelofsideformuladeltaplusnegexistsdelta[1]
{\varilesskernelofsideformuladeltaplusnegexistsdelta{#1}\sforallvari}

\newcommand\varilesskernelofsideformuladeltaplusnegexistsdelta[2]
{0\tight<#2\delta{#1}\nottight{\und}
 \secondhalfvarilesskernelofsideformuladeltaplusnegexistsdelta{#1}{#2}}

\newcommand\secondhalfvarilesskernelofsideformuladeltaplusnegexistsdelta[2]
{\forall\boundvari x{#1}\boldunequal\wforallvari x 0\stopq
\twicesecondhalfvarilesskernelofsideformuladeltaplusnegexistsdelta
{#1}{#2}\boundvari}

\newcommand\twicesecondhalfvarilesskernelofsideformuladeltaplusnegexistsdelta[3]
{\inparentheses{
        |\app{\wforallvari{#1}{}}{#3 x{#1}}\tight-\wforallvari y{#1}|
        <\existsvari\varepsilon{#1}
      \nottight\antiimplies
        |#3 x{#1}\tight-\wforallvari x 0|<#2\delta{#1}}}

\newcommand\varilesssideformuladeltaplusnegexistsdelta[2]{\neg\inparentheses
{\varilesskernelofsideformuladeltaplusnegexistsdelta{#1}{#2}}}

\newcommand\vcdeltaplusnegexistsdelta[1]{\{\wforallvari x 0,\wforallvari
{#1}{},\wforallvari y{#1}, \existsvari\varepsilon {#1}\}\times\{\sforallvari
\delta {#1}\}}

\newcommand\figurefsix{\figuref{figure six}}

\newcommand\criticalbetaformula{0\tight<\existsvari\delta{}
    \und
    \forall x\boldunequal\wforallvari x 0\stopq 
    \inparenthesesoplist{
        |\inpit{\app{\wforallvari f{}}{x}\tight+\app{\wforallvari g{}}{x}}
         \tight-\inpit{\wforallvari y f\tight+\wforallvari y g}|
        <\wforallvari\varepsilon{}
      \oplistantiimplies
        |x\tight-\wforallvari x 0|<\existsvari\delta{}}}

\newcommand\protofifthcase{\begin{array}[t]{@{}l@{}}
  \neg\inparentheses{
\oplistexpandedsecondhalfvarilesskernelofsideformuladeltaplusnegexistsdelta
f\sforallvari}\comma
\\\neg\inparentheses{
\oplistexpandedsecondhalfvarilesskernelofsideformuladeltaplusnegexistsdelta
g\sforallvari}\comma
\\\lastlineoffifthcase
\\\end{array}}

\newcommand\fifthcase{\math{
|\app{\wforallvari f{}}{\existsvari x f}-\wforallvari y f|
\nless\existsvari\varepsilon f\comma
|\app{\wforallvari g{}}{\existsvari x g}-\wforallvari y g|
\nless\existsvari\varepsilon g\comma
}\nopagebreak\\\mbox{}\hfill\lastlineoffifthcase}

\newcommand\lastlineoffifthcase{\math{\sforallvari x{}\boldequal\wforallvari x 0
  \comma t<\wforallvari\varepsilon{}
  \comma  |\sforallvari x{}\tight-\wforallvari x 0|
         \nless\existsvari\delta{}
  \comma\Omega}}

\newcommand
\oplisttwicesecondhalfvarilesskernelofsideformuladeltaplusnegexistsdelta[3]
{\inparenthesesoplist{
        |\app{\wforallvari{#1}{}}{#3 x{#1}}
         \tight-\wforallvari y{#1}|<\existsvari\varepsilon{#1}
      \oplistantiimplies
        |#3 x{#1}\tight-\wforallvari x 0|<#2\delta{#1}}}

\newcommand\oplistsecondhalfvarilesskernelofsideformuladeltaplusnegexistsdelta
[2]{\forall\boundvari x{#1}\boldunequal\wforallvari x 0\stopq
\oplisttwicesecondhalfvarilesskernelofsideformuladeltaplusnegexistsdelta
{#1}{#2}\boundvari}

\newcommand
\oplistexpandedsecondhalfvarilesskernelofsideformuladeltaplusnegexistsdelta[2]
{\existsvari x{#1}\boldunequal\wforallvari x 0\nottight\implies
\oplisttwicesecondhalfvarilesskernelofsideformuladeltaplusnegexistsdelta
{#1}{#2}\existsvari}

\newcommand\oplistvarilesskernelofsideformuladeltaplusnegexistsdelta[2]
{0\tight<#2\delta{#1}\nottight{\und}
 \oplistsecondhalfvarilesskernelofsideformuladeltaplusnegexistsdelta{#1}{#2}}

\newcommand\oplistkernelofsideformuladeltaplusnegexistsdelta[1]
{\oplistvarilesskernelofsideformuladeltaplusnegexistsdelta{#1}\sforallvari}

\newcommand\oplistprincipalformuladeltaplusnegexistsdelta[1]
{\neg\exists\boundvari\delta #1\stopq\inparentheses
{\oplistvarilesskernelofsideformuladeltaplusnegexistsdelta{#1}\boundvari}}

\newcommand\oplistvarilesssideformuladeltaplusnegexistsdelta
[2]{\neg\inparentheses
{\oplistvarilesskernelofsideformuladeltaplusnegexistsdelta{#1}{#2}}}

\newcommand\oplistccdeltaplusnegexistsdelta[1]
{{\sforallvari\delta {#1}}\mapsto\inparentheses{
   {\oplistkernelofsideformuladeltaplusnegexistsdelta{#1}}}}

\newcommand\figuredeltaminusinstead[1]{\begin{figure}[#1]\LINEmath{\xymatrix{
&{\existsvari\varepsilon f}\ar[ddr]\ar@/^/[ddrrr]\ar@/^/[drr]
&&&&{\existsvari\varepsilon g}\ar[ddl]\ar@/_/[ddlll]\ar@/_/[dll]&
\\&&{\existsvari x f}&{\wforallvari x{}}\ar@/_/[l]\ar@/^/[r]&{\existsvari x g}
\\&&{\wforallvari\delta f}\ar[r]
&{\existsvari\delta{}}\ar[u]\ar@/^/[l]\ar@/_/[r]
&{\wforallvari\delta g}\ar[l]
}}\caption{\protect\begin{tabular}[t]{@{}l@{}}(Cyclic) State of 
\vc\ \protect\nlbmath R 
\protect\\for alternative proof of \protect\sectref{section minus}
with \deltaminus-rules only\protect\end{tabular}}\label
{figure delta minus}\end{figure}}

\newcommand\figurefinalvc[1]{\label
{section where variable condition after sigma is}\begin{figure}[#1]
{\begin{minipage}{\textwidthminustwomm}\LINEmath{\xymatrix{
&{\existsvari\varepsilon f}\ar[ddr]
&&{\wforallvari\varepsilon{}}\ar[d]\ar@{.>}@/_/[ll]\ar@{.>}@/^/[rr]
&&{\existsvari\varepsilon g}\ar[ddl]
\\&&{\existsvari x f}&{\sforallvari x{}}\ar@/_/[l]\ar@/^/[r]&{\existsvari x g}
\\{\wforallvari y f}\ar@/_/[urrr]\ar[rr]
&&{\sforallvari\delta f}\ar[r]&{\existsvari\delta{}}\ar[u]
&{\sforallvari\delta g}\ar[l]
&&{\wforallvari y g}\ar@/^/[ulll]\ar[ll]
\\&&{\wforallvari f{}}\ar[u]\ar[uur]
&{\wforallvari x 0}\ar[ur]\ar[ul]\ar@/_/[uu]
&{\wforallvari g{}}\ar[u]\ar[uul]
}}\end{minipage}}
\caption{\protect\begin{tabular}[t]{@{}l@{}}
 (Acyclic) \VC\ 
 \protect\nlbmath R.\protect\\With dotted edges: Final State
 in \protect\sectref{section clean up}.\protect\\
 Without dotted edges:\protect\\
 State after application of \protect\nlbmath{\protect\sigma}, both in 
 \protect\sectref{section partial success} and in
 \protect\sectref{section backtracking}\protect\\\protect\end{tabular}}\label
{figure variable-condition final}\label
{figure variable-condition one}\vspace*{-6ex}\end{figure}}

\newcommand\figurevision[1]{\begin{figure}[#1]\fbox{\begin
{minipage}{\textwidthminustwomm}
\LINEmath{\xymatrix{
 &{(1^5.3.1^2)}
  \ar[dl]^{\beta_1}
  \ar[d]^{\beta_2}
\\{{\framebox{\framebox{\math{
      \existsvari x f\boldunequal\wforallvari x 0}}}}
  \comma B\comma C\comma\Omega}
 &{A\comma B\comma C\comma\Omega}
  \ar[dl]^{\beta_1}
  \ar[d]^{\beta_2}
\\{A\comma B\comma 0\tight<\existsvari\delta{}\comma\Omega}
 &{A\comma B\comma \forall x\boldunequal\wforallvari x 0\stopq 
   \inparenthesesoplist{
        |\ldots
         \tight-\inpit{\wforallvari y f\tight+\wforallvari y g}|
        <\wforallvari\varepsilon{}
      \oplistantiimplies
        |x\tight-\wforallvari x 0|<\existsvari\delta{}}\comma\Omega}
  \ar[d]^{\forall,\,\,\app{\delta_0^+}{\sforallvari x{}},\,\,\alpha_0^2}
\\
 &{\begin{array}[b]{@{}c@{}}
   A\comma B\comma \framebox{{\framebox{\math{
       \sforallvari x{}\boldequal\wforallvari x 0}}}}\comma
   {\framebox{\math{|\sforallvari x{}
            \tight-\wforallvari x 0|\nless\existsvari\delta{}}}}\comma
   \\t<\wforallvari\varepsilon{}
   \comma\Omega
   \end{array}}
}}\par\noindent\mbox{}
\par\yestop\noindent
Here \math A denotes the formula
\bigmaths{\neg\inparentheses{|\app{\wforallvari f{}}{\existsvari x f}
    \tight-\wforallvari y f|<\existsvari\varepsilon f
      \nottight\antiimplies
      \framebox{\math{
        |\existsvari x f\tight-\wforallvari x 0|<\sforallvari\delta f}}}}.
\math{B} and \math{C} denote the second and third \math\beta-formula of the
sequent \math{(1^5.3.1^2)}, respectively.
And \math{\Pi} the sequent at the second (\math{\beta_2}-) 
child of the root without the second \math\beta-formula, 
\ie\ without the third \math\beta-formula of \nlbmath{(1^5.3.1^2)}.%
\end{minipage}}
\caption{\label{figure vision}\protect\begin{tabular}[t]{@{}l@{}}
\NonPermutability\ of \protect\math\beta\ 
at \protect\nlbmath{(1^5.3.1^2)} and\protect\\ 
\protect\math{\beta} at the \protect\math{\beta_2}-child of 
\protect\nlbmath{(1^5.3.1^2)}: \protect\\
No chance to prove \protect\math{\protect\existsvari x f\protect\boldunequal
\protect\wforallvari x 0} at leftmost leaf\protect\end{tabular}}
\end{figure}}

\newcommand\figurefailedproof[1]{\begin{figure}[#1]\LINEmath{\xymatrix{
 &{(1)}
  \ar[d]_{1^2}^{
    \alpha_0^2,\,\,\lim,\,\,\forall,\,\,
    \app{\delta_0^-}{\wforallvari\varepsilon{}},\,\,\alpha_0,\,\,\exists}
\\&(1^3)\ar[d]_{1}^{
      \app{\gamma_0}{\min(\sforallvari\delta f,\sforallvari\delta g)}}
\\&(1^4)\ar[d]_{1}^{\lim^2,\,\,\forall^2,\,\,
      \app{\gamma_0}{\existsvari\varepsilon f},\,\,
      \app{\gamma_0}{\existsvari\varepsilon g}}
\\&(1^5)
 \ar[dl]_{1}^{\beta_1}\ar[d]_{2}^{\beta_2,\,\,\beta_1}
 \ar[dr]_{3}^{\beta_2^2,\,\,\exists^2}
\\{(1^5.1)}&{(1^5.2)}
 &{(1^5.3)}\ar[d]_{1}^{\app{\delta_0^+}{\sforallvari\delta f}}
\\&&{(1^5.3.1)}\ar[dl]_{1}^{\beta_1}\ar[d]_{2}^{\beta_2}
\\&{(1^5.3.1.1)}
 &{(1^5.3.1.2)}
  \ar[d]_{1}^{\app{\delta_0^+}{\sforallvari\delta g},\,\,\alpha_0^2,\,\,\forall}
\\&&{(1^5.3.1.2.1)}
  \ar[d]_{1}^{\app{\delta_0^+}{\sforallvari x{}},\,\,\alpha_0^2}
\\&&{(1^5.3.1.2.1^2)}
  \ar[d]_{1}^{\forall^2,\,\,\app{\gamma_0}{\sforallvari x{}}^2}
\\&&{(1^5.3.1.2.1^3)}
}}\caption
{\label{figure failure}\protect\begin{tabular}[t]{@{}l@{}}
\NonPermutability\ of \protect\math\beta\ 
at \protect\nlbmath{(1^5.3.1)} and 
\protect\deltaplus\ at \protect\nlbmath{(1^5.3.1.2)}: 
\protect\\
No chance to prove 
\protect\math{0\protect\tight<\protect\min
(\protect\sforallvari{\protect\delta}f,\protect\sforallvari
{\protect\delta}g)} at \protect\nlbmath{(1^5.3.1.1)}
\protect\end{tabular}}
\end{figure}}

\def\mycaptionfigure{%
     \refstepcounter{figure}
     \expandafter\@firstofone
   {\@dblarg{\@caption{figure}}}%
}

\newcommand\figureglobaldefinitions[1]{\begin{figure}[#1]\fbox
{\begin{minipage}{\textwidthminustwomm}
In the proof below, 
(2), (3), (4), (5), (6), (7), (8), (9) (where the boxes around the formulas 
just indicate the matching in the lemma application) and
\math\Gamma, \math\Xi, \math\Theta, \math\Omega\ and \nlbmath\sigma\ 
and \nlbmath t abbreviate the 
following lemmas and sequents and substitution and term, respectively:
\par\halftop\noindent\math{\begin{array}{@{}l l@{}}
  (2)\mbox:
 &\min(\wforallvari y{},\wforallvari z{})\leq\wforallvari y{}
\footroom\\(3)\mbox:
 &\wforallvari z 4\tight<\wforallvari z 6\comma
  \wforallvari z 4\tight\nless\wforallvari z 5\comma
  \wforallvari z 5\tight\nleq\wforallvari z 6
\footroom\\(4)\mbox:
 &\wforallvari z 9\tight<\min(\wforallvari z{10},\wforallvari z{11})\comma
   \wforallvari z 9\tight\nless\wforallvari z{10}\comma
   \wforallvari z 9\tight\nless\wforallvari z{11}
\footroom\\(5)\mbox:
 &
  |\inpit{\wforallvari z 0\tight+\wforallvari z 1}
   \tight-
   \inpit{\wforallvari z 2\tight+\wforallvari z 3}|
  \leq
  |\wforallvari z 0\tight-\wforallvari z 2|
  +
  |\wforallvari z 1\tight-\wforallvari z 3|
\footroom\\(6)\mbox:
 & \framebox{\wforallvari z 4\tight<\wforallvari z 6}
 \comma
   \framebox{\framebox{
   \wforallvari z 4\tight\nleq\wforallvari z 5
   }}
 \comma
   \wforallvari z 5\tight\nless\wforallvari z 6
\footroom\\(7)\mbox:
 &  \wforallvari z{12}\tight+\wforallvari z{13}
    \tight<
    \wforallvari z{14}\tight+\wforallvari z{15}
  \comma
    \framebox{\math{
    \wforallvari z{12}
    \tight\nless
    \wforallvari z{14}
    }}
  \comma\framebox{\framebox{\math{
    \wforallvari z{13}
    \tight\nless
    \wforallvari z{15}}}}
\footroom\\(8)\mbox:
 &{{\wforallvari\varepsilon{}\over 2}\tight+
   {\wforallvari\varepsilon{}\over 2}}
  \leq
  \wforallvari\varepsilon{}
\footroom\\(9)\mbox:
 &0\tight<{\wforallvari\varepsilon{}\over 2}\comma 
  0\tight\nless\wforallvari\varepsilon{}
\\\end{array}}
\par\yestop\noindent{\Huge\math{\Gamma}:} 
\hfill\math{\begin{array}[t]{@{}r@{}}
    \neg\forall\varepsilon_f\stopq\inparentheses{
    0\tight<\varepsilon_f
    \implies
    \exists\delta_f\tight>0\stopq 
    \forall x_f\boldunequal\wforallvari x 0\stopq 
    \inparenthesesoplist{
        |\app{\wforallvari f{}}{x_f}-\wforallvari y f|<\varepsilon_f
      \oplistantiimplies
        |x_f\tight-\wforallvari x 0|<\delta_f
    }}\comma
  \\\neg\forall\varepsilon_g\stopq\inparentheses{
    0\tight<\varepsilon_g
    \implies
    \exists\delta_g\tight>0\stopq 
    \forall x_g\boldunequal\wforallvari x 0\stopq 
    \inparenthesesoplist{
        |\app{\wforallvari g{}}{x_g}-\wforallvari y g|<\varepsilon_g
      \oplistantiimplies
        |x_g\tight-\wforallvari x 0|<\delta_g
    }}\comma
  \\\exists\delta\stopq\inparentheses{
    0\tight<\delta
    \und
    \forall x\boldunequal\wforallvari x 0\stopq 
    \inparenthesesoplist{
        |\inpit{\app{\wforallvari f{}}{x}\tight+\app{\wforallvari g{}}{x}}
         -\inpit{\wforallvari y f\tight+\wforallvari y g}|
         <\wforallvari\varepsilon{}
      \oplistantiimplies
        |x\tight-\wforallvari x 0|<\delta}}
  \\\end{array}}
\par\yestop\yestop\noindent{\Huge\math{\Xi}:} 
\hfill\math{\begin{array}[t]{@{}r@{}}
\criticalbetaformula\comma
\\0\tight\nless\wforallvari\varepsilon{}\comma\Gamma
\\\end{array}}
\par\yestop\noindent{\Huge\math{\Theta}:} 
\hfill\math{\begin{array}[t]{@{}r@{}}
\oplistvarilesssideformuladeltaplusnegexistsdelta f\sforallvari\comma
\\\oplistprincipalformuladeltaplusnegexistsdelta g\comma
\\0\tight\nless\wforallvari\varepsilon{}\comma\Gamma
\\\end{array}}
\par\yestop\noindent{\Huge\math{\Omega}:} 
\hfill\math{\begin{array}[t]{@{}r@{}}
0\tight\nless\sforallvari\delta f
\comma
\neg\oplistsecondhalfvarilesskernelofsideformuladeltaplusnegexistsdelta 
f\sforallvari
\comma
\\0\tight\nless\sforallvari\delta g
\comma
\neg\oplistsecondhalfvarilesskernelofsideformuladeltaplusnegexistsdelta 
g\sforallvari\comma
\\0\tight\nless\wforallvari\varepsilon{}\comma\Gamma
\\\end{array}}
\par\noindent{\Huge\math{\sigma}:} 
\LINEmath{\begin{array}[t]{@{}r@{}}
  \{
  \existsvari x f\tight\mapsto\sforallvari x{}\comma
  \existsvari x g\tight\mapsto\sforallvari x{}\comma
  \existsvari\delta{}\tight\mapsto
    \min(\sforallvari\delta f,\sforallvari\delta g)
  \}
\\\end{array}}%
\par\halftop\noindent{\Huge\math t:} 
\LINEmath{\begin{array}[t]{@{}r@{}}
 |\,\,\inpit{\app{\wforallvari f{}}{\sforallvari x{}}
   \tight+\app{\wforallvari g{}}{\sforallvari x{}}}
   -\inpit{\wforallvari y f\tight+\wforallvari y g}\,\,|
\\\end{array}}%
\end{minipage}}\caption{Global abbreviations for the proof of 
\sectref{section proof}}\label
{figure global abbreviations}\vspace*{-16ex}\end{figure}}

\newcommand\thereductiverules[1]{\begin{figure}[#1]\fbox
{\begin{minipage}{\textwidthminustwomm}
Let \math{A} and \math{B} be formulas.
Let \math{\Gamma} and \math{\Pi} 
be sequents, \ie\ 
disjunctive lists of formulas.
Let \math{x\in\Vbound} be a bound variable,
and let \math{\mathcal F} be the current proof forest,
such that \VAR{{\mathcal F}} \nolinebreak contains all variables
already in use, especially those from \math\Gamma, \math\Pi, and \math A. 
Note that \overline A is the\emph{conjugate} of the formula \nlbmath A, \ 
\ie\ \nlbmath B if \nlbmath A is of the form\ \nlbmath{\neg B}, \ 
and \nlbmath{\neg A} otherwise. 
\par\yestop\noindent{\bf\math\alpha-rules \math{\alpha\over\alpha_0}: } \ 
\shortstrongexpansionrule{\Gamma~\neg\neg A~\Pi}{A~\Gamma~\Pi}{}{}{}{}
\shortstrongexpansionrule{\Gamma~\inpit{A\tightoder B}~\Pi}{A~B~\Gamma~\Pi}{}{}{}{}
\shortstrongexpansionrule{\Gamma~\neg\inpit{A\tightund B}~\Pi}{\overline{\,A\,}~\overline{\,B\,}~\Gamma~\Pi}{}{}{}{}
\shortstrongexpansionrule{\Gamma~\inpit{A\tightimplies B}~\Pi}{\overline{\,A\,}~B~\Gamma~\Pi}{}{}{}{}
\shortstrongexpansionrule{\Gamma~\inpit{A\tightantiimplies B}~\Pi}{A~\overline{\,B\,}~\Gamma~\Pi}{}{}{}{}
\par\yestop\halftop\noindent
{\bf\math\beta-rules \math{\beta\over{\beta_1\rule{0ex}{1.1ex}\atop\beta_2}}: }
\branchingstrongexpansionrule
{\Gamma~\inpit{A\tightund B}~\Pi}
{A~
\Gamma~\Pi}
{}{}{}{}
{B~
\Gamma~\Pi}
\branchingstrongexpansionrule
{\Gamma~\neg\inpit{A\tightoder B}~\Pi}
{\overline{\,A\,}~
\Gamma~\Pi}
{}{}{}{}
{\overline{\,B\,}~
\Gamma~\Pi}
\branchingstrongexpansionrule
{\Gamma~\neg\inpit{A\tightimplies B}~\Pi}
{A~
\Gamma~\Pi}
{}{}{}{}
{\overline{\,B\,}~
\Gamma~\Pi}
\branchingstrongexpansionrule
{\Gamma~\neg\inpit{A\tightantiimplies B}~\Pi}
{\overline{\,A\,}~
\Gamma~\Pi}
{}{}{}{}
{B~
\Gamma~\Pi}
\par\halftop\yestop\noindent{\bf\math\gamma-rules \math{\gamma\over\app{\gamma_0}t}: }
Let \math{t} be any term (by default a new \fev):
\par\halftop\noindent\mbox{}
\hfill\shortstrongexpansionrule
{\hfill\Gamma~~~\exists x.A~~~\Pi}
{A\{x\tight\mapsto t\}~~~\Gamma~~~\exists x.A~~~\Pi}
{}{}{}{}\hfill
\shortstrongexpansionrule
{\hfill\Gamma~~~\neg\forall x.A~~~\Pi}
{\overline{\,A\{x\tight\mapsto t\}\,}~~~\Gamma
 ~~~\neg\forall x.A~~~\Pi}
{}{}{}{}
\par\halftop\yestop\noindent{\bf\deltaminus-rules 
\math{\delta\over\app{\delta_0^-}{\wforallvari x{}}}: }
Let \bigmath{\wforallvari x{}\nottight\in\Vwall\setminus\VAR{\mathcal F}}
be a new \wfuv:\par\noindent
\strongexpansionrule
{\Gamma~~~\forall x.A~~~\Pi}
{A\{x\tight\mapsto\wforallvari x{}\}~~~\Gamma~~~\Pi}
{}
{\VARsomesall{\Gamma~~\forall x.A~~\Pi}\times\{\wforallvari x{}\}}
{}{}
\par\halftop\noindent
\strongexpansionrule
{\Gamma~~~\neg\exists x.A~~~\Pi}
{\majorheadroom
 \overline{\,A\{x\tight\mapsto\wforallvari x{}\}\,}~~~\Gamma~~~\Pi}
{}
{\VARsomesall{\Gamma~~\neg\exists x.A~~\Pi}\times\{\wforallvari x{}\}}
{}{}
\par\yestop\yestop\noindent{\bf\deltaplus-rules
\math{\delta\over\app{\delta_0^+}{\sforallvari x{}}}: }
Let \bigmath{\sforallvari x{}\nottight\in\Vsall\setminus\VAR{\mathcal F}}
be a new \sfuv:\par\noindent
\strongexpansionrule
{\Gamma~~~\forall x.A~~~\Pi}
{A\{x\tight\mapsto\sforallvari x{}\}~~~\Gamma~~~\Pi}
{\{\pair
     {\sforallvari x{}}
     {\headroom\overline{\,A\{x\tight\mapsto\sforallvari x{}\}\,}}
\}}
{\VARfree{\forall x.A}
 \times
 \{\sforallvari x{}\}
}
{\revrelapp{\tight\leq}{\VARall A}
 \times
 \{\sforallvari x{}\}
}{}
\par\yestop\yestop\noindent
\strongexpansionrule
{\Gamma~~~\neg\exists x.A~~~\Pi}
{\majorheadroom\overline{\,A\{x\tight\mapsto\sforallvari x{}\}\,}~~~\Gamma
 ~~~\Pi}
{\{\pair
     {\sforallvari x{}}
     {A\{x\tight\mapsto\sforallvari x{}\}}
\}}
{\VARfree{\neg\exists x.A}
 \times
 \{\sforallvari x{}\}
}
{\revrelapp{\tight\leq}{\VARall A}
 \times
 \{\sforallvari x{}\}
}{}\end{minipage}}\caption
{The reductive rules of our calculus}\label{figure reductive rules}\end{figure}}

\newcommand\figurecompleteproof[1]{%
\begin{figure}[#1]\noindent\mbox{}\hfill\LINEmath
{\xymatrix{
 &{(1)}
  \ar[d]_{1^2}^{
    \alpha_0^2,\,\,\lim,\,\,\forall,\,\,
    \app{\delta_0^-}{\wforallvari\varepsilon{}},\,\,\alpha_0,\,\,\exists}
&&\mbox{\underline{~~~~~~\sectref{section explanation and initialization}~~~~~~}}
\\
 &(1^3)
  \ar[d]_{1}^{\app{\gamma_0}{\min(\sforallvari\delta f,\sforallvari\delta g)}}
&&\mbox{{~~~~~~\sectref{section expanding the proof}~~~~~~}}
\\&(1^4)\ar[d]_{1}^{\lim^2,\,\,\forall^2,\,\,
        {\gamma_0}
        \left(\begin{array}{@{}c@{}}{\wforallvari\varepsilon{}}\over 2
        \\\end{array}\right)
      ^2}
&&\mbox{{~~~~~~\vdots~~~~~~}}
\\&(1^5)
 \ar[dl]_{1}^{\beta_1}\ar[d]_{2}^{\beta_2,\,\,\beta_1}
 \ar[dr]_{3}^{\beta_2^2,\,\,\exists^2}
&&\mbox{{~~~~~~\vdots~~~~~~}}
\\{(1^5.1)}
  \ar[d]_{1}^{\mbox{\rm\tiny lemma (9)}}
 &{(1^5.2)}
  \ar[d]_{1}^{\mbox{\rm\tiny lemma (9)}}
 &{(1^5.3)}\ar[d]_{1}^{\app{\delta_0^+}{\sforallvari\delta f}}
 &\mbox{\underline{~~~~~~\sectref{section expanding the proof}~~~~~~}}
\\{\bullet}
 &{\bullet}
 &{\mathbf{(1^5.3.1)}}
  \ar[d]_{\mathbf{1'}}^{\mathbf{\app{\delta_0^+}{\sforallvari\delta g}}}
 &{\mbox{\bf non-}}
\\
 &
 &{\mathbf{(1^5.3.1.1')}}
  \ar[dl]_{\mathbf{1}}^{\mathbf{\beta_1}}
  \ar[d] _{\mathbf{2}}^{\mathbf{\beta_2}}
 &{\mbox{\bf permutable}}
\\
 &{\mathbf{(1^5.3.1.1'.1)}}
  \ar[d]_{1}^{\mbox{\rm\tiny lemma (4)}}
 &{\mathbf{(1^5.3.1.1'.2)}}
  \ar[d]_{\rm copy}
  ^{\alpha_0^2,\,\,\app{\gamma_0}{\sforallvari x{}}^2,\,\,\forall^3}
 &\mbox{\underline{\bf~~~~~~~steps~~~~~~~~}}
\\
 &{\bullet}
 &{(1^5.3.1^2.2)}
  \ar[d]_{1}^{\app{\delta_0^+}{\sforallvari x{}},\,\,\alpha_0^2}
 &\mbox{{~~~~~~\sectref{section backtracking}~~~~~~}}
\\&&{(1^5.3.1^2.2.1)}
  \ar[dll]_{1}^{\beta_1}
  \ar[dl]_{2}^{\beta_2,\,\,\beta_1}
  \ar[d]_{}^{\beta_2^2}
 &\mbox{{~~~~~~\vdots~~~~~~}}
\\{(1^5.3.1^2.2.1.1)}
 &{(1^5.3.1^2.2.1.2)}
 &{\bullet}
  \ar[dll]_{3}^{\beta_1}
  \ar[dl]_{4}^{\beta_2,\,\,\beta_1}
  \ar[d]_{5}^{\beta_2^2}
 &\mbox{{~~~~~~\vdots~~~~~~}}
\\{(1^5.3.1^2.2.1.3)}
  \ar[d]_{1}^{\mbox{\rm\tiny lemma (2,3)}}
 &{(1^5.3.1^2.2.1.4)}
  \ar[d]_{1}^{\mbox{\rm\tiny lemma (2,3)}}
 &{(1^5.3.1^2.2.1.5)}
  \ar[d]_{1}^{\mbox{\rm\tiny lemma (5)}}
 &\mbox{\underline{~~~~~~\sectref{section backtracking}~~~~~~}}
\\{\bullet}
 &{\bullet}
 &{(1^5.3.1^2.2.1.5.1)}
  \ar[d]_{1.1.1}^{\mbox{\rm\tiny lemma (6,7,8)}}
 &\mbox{\underline{~~~~~~\sectref{section immediate focus}~~~~~~}}
\\&&{\bullet}
 &\mbox{\underline{~~~~~~\sectref{section clean up}~~~~~~}}}}%
\caption
{\label{figure six}Closed proof tree with 
\protect\nonpermutable\ \protect\nlbmath{\protect\beta}
and \protect\deltaplus-step}\vspace*{-15ex}\end{figure}\pagebreak}

\input seki-deckblatt

\newlength{\sectionintocsep} %separation for section line in table of contents
\newlength{\contentsandreferencesheadroom} %room over line ``Contents''
\newlength{\contentsandreferencesfootroom} %room under line ``Contents''
\catcode`\@=11

\def\tableofcontents{\ignorespaces
\section*{\contentsname\@mkboth
{\uppercase{\contentsname}}{\uppercase{\contentsname}}}%
\vspace*\contentsandreferencesfootroom
\@starttoc{toc}}
\def\contentsname{Contents} % <----------

\catcode`\@=12

\input headerhot

\newcommand\Proofof{Proof of}
\renewenvironment{proof}[1]{\yestop\begin{sloppypar}\noindent
{\bf\Proofof\ {#1}}\\}{\end{sloppypar}}
\renewenvironment{proofqed}[1]
{\begin{sloppypar}\def\fooqed{#1}\noindent{\bf\Proofof\ \fooqed}}
{\QEDbf\fooqed\end{sloppypar}}
\renewenvironment{proofparsep}[1]{\parindent=0pt\begin
{sloppypar}\noindent{\bf\Proofof\ {#1}}\nopagebreak\par}
{\end{sloppypar}}
\renewenvironment{proofparsepqed}[1]{\parindent=0pt\begin
{sloppypar}\def\fooqed{#1}\noindent{\bf\Proofof\ \fooqed}\nopagebreak\par}
{\nopagebreak\QEDbf\fooqed\end{sloppypar}}
\renewenvironment{proofshort}[1]{\begin{sloppypar}\noindent
{\bf\Proofof\ {#1}:}}{\end{sloppypar}}

{\end{enumerate}\renewcommand{\theenumi}{\arabic{enumi}}}

\mathcommand\myfootnotemark[1]{^{#1}}

\newcommand\repname{{\rm set}}
\mathapplycommand\rep\repname
\mathcommand\repr[1]{{\repname[{#1}]}}
\mathcommand\msa{\langle}
\mathcommand\mse{\rangle}
\mathcommand\msu{\,\sqcup\,}
\mathcommand\msin{{\rm\;in\;}}
\mathcommand\mssetminus{\setminus\!\!\setminus}
\mathcommand\tightmssubseteq{\sqsubseteq}
\mathcommand\mssubseteq{\ \tightmssubseteq\ }
\mathcommand\approxapprox{\approx\:\!\!\approx}
\mathcommand\quasilquasil{\,\lesssim\!\lesssim\,}
\mathcommand\quasibquasib{\,\gtrsim\!\gtrsim\,}
\mathcommand\fmul[1]{{\rm FMul}(#1)}
\mathcommand\smul[1]{{\rm SMul}(#1)}
\mathcommand\multisetwith [2]{\msa\ {#1}\ |\ {#2}\ \mse}
\mathcommand\multisetwithq[3]{\msa\ {#2}\ |_{#1}\ {#3}\ \mse}
\newcommand\superseteq     {\supseteq}

\mathcommand\quasilhd
{\mathop{\mbox{\raisebox{0.31ex}{\math\lhd}\hspace{-0.75em}\raisebox
{-0.6ex}{\math\sim}}}}
\newcommand\quasirhd{\mbox{\raisebox{0.31ex}{$\rhd$}\hspace{-0.75em}\raisebox{-0.6ex}{$\sim$}}}

\mathcommand\rhdrhd{\rhd$\hspace{-0.35em}$\rhd}
\mathcommand\lhdlhd{\lhd$\hspace{-0.21em}$\lhd}

\mathcommand\quasilhdquasilhd{\quasilhd$\hspace{-0.13em}$\quasilhd}

\newcommand\hiddensubSS{_{_{\rm SS}}}
\mathcommand\antisubsum     {\rhd\hiddensubSS}
\mathcommand\notantisubsum  {\ntriangleright\hiddensubSS}
\mathcommand\subsum         {\lhd\hiddensubSS}
\mathcommand\notsubsum      {\ntriangleleft\hiddensubSS}
\mathcommand\antisubsumeq   {\trianglerighteq\hiddensubSS}
\mathcommand\subsumeq       {\trianglelefteq\hiddensubSS}
\mathcommand\quasisubsum    {\,\quasilhd\raisebox{0.1ex}{$\hiddensubSS$}}
\mathcommand\antiquasisubsum{\,\quasirhd\raisebox{0.1ex}{$\hiddensubSS$}}
\mathcommand\quasiquasisubsum{\quasisubsum\!\!\quasisubsum}
\mathcommand\antiquasiquasisubsum{\antiquasisubsum\!\!\!\antiquasisubsum}

\newcommand\hiddensubH{_{_{\rm H}}}
\newcommand\hiddensubCONS{_{_\CONS}}
\mathcommand\hql   {\,\lesssim\hiddensubH}
\mathcommand\consql{\,\lesssim\hiddensubCONS}
\mathcommand\hl    {\,<       \hiddensubH}
\mathcommand\hleq  {\,\leq    \hiddensubH}
\mathcommand\consl {\,<       \hiddensubCONS}
\mathcommand\conseq{\,\approx \hiddensubCONS}

\newcommand\inthesequel{in what follows}
\newcommand\cons {{\rm cons}}
\mathcommand\sigconsV{\sig/\cons/\V}
\mathcommand\sigconsR{\sig/\cons/\R}
\mathcommand\primesigconsV{\sig'\!/\cons'\!/\V'}
\mathcommand\primesigconsR{\sig'\!/\cons'\!/\R'}
\newcommand\dunno{{\rm sc}}
\newcommand\DUNNO{{\rm SC}}
\mathcommand\SIGCONS   {\{\SIG,\CONS\}}
\mathcommand\sigsortstimes{\SIGCONS\tight\times\sigsorts}

\mathcommand\TERMSSYM
{\mathcal{T\hspace{-.20em}E\hspace{-.08em}R\hspace{-.16em}M\hspace{-.10em}S}}
\mathapplycommand\condterms{\TERMSSYM}
\mathcommand\kurzregel{((l,r),C)}
\mathcommand\kurzregelprime{((l',r'),C')}
\mathcommand\kurzregelindex[1]{((l_{#1},r_{#1}),C_{#1})}
\mathapplycommand\lhs{\rm lhs}

\mathcommand\red{\redsimple} %conflicts with ps tricks

\mathcommand\lemms{L}
\mathcommand\hypos{H}
\mathcommand\goals{G}
\mathcommand\lemmsprime{\lemms'}
\mathcommand\hyposprime{\hypos'}
\mathcommand\goalsprime{\goals'}
\mathcommand\lemmsprimeprime{\lemms''}
\mathcommand\hyposprimeprime{\hypos''}
\mathcommand\goalsprimeprime{\goals''}
\mathcommand\oldtriple            {(\lemms   ,\hypos   ,\goals  )}
\mathcommand\inittriple        {(\emptyset,\emptyset,\goals  )}
\mathcommand\triplehelp[1]     {(\lemms#1,\hypos#1 ,\goals#1)}
\mathcommand\tripleprime       {\triplehelp'}
\mathcommand\triplenogoalsprime{(\lemmsprime,\hyposprime,\emptyset  )}
\mathcommand\tripleprimeprime  {\triplehelp{''}}
\mathcommand\tripleindex[1]    {\triplehelp{_{#1}}}

\mathcommand\constcong[1]{\,\,\sim_{\!_{#1}}\,}

\mathapplycommand\avail{\rm\Av ail}

\def\emph#1{\/ {\itshape#1}\/}
\newcommand\tightemph[1]{\/{\itshape#1}\/}

\input headertheorem
\mathcommand\ident[1]{\mathsf{#1}}
\newcommand\plussymbol  {\ident{+}}
\newcommand\minussymbol {\ident{-}}
\newcommand\dividesymbol{\ident{/}}
\newcommand\timessymbol {\ident{*}}

\newcommand\set     {\ident{set}}

\newcommand\naturalssymbol{\ident{naturals}}
\newcommand\gensymsymbol{\ident{gensym}}
\mathcommand\mbpsymbol{\ident{m\hspace{-0.055em}b\hspace{-0.045em}p}}

\newcommand\csymbol     {\ident c}
\newcommand\esymbol     {\ident e}
\newcommand\fsymbol     {\ident f}
\newcommand\gsymbol     {\ident g}
\newcommand\hsymbol     {\ident h}
\newcommand\ksymbol     {\ident k}
\newcommand\psymbol     {\ident p}
\newcommand\ssymbol     {\ident s}
\newcommand\Everysymbol {\ident{Every}}
\newcommand\Permsymbol {\ident{Perm}}
\newcommand\RExistssymbol{\ident{Rexists}}
\newcommand\invertsymbol{\ident{invert}}
\newcommand\invsymbol{\ident{inv}}
\newcommand\abssymbol   {\ident{abs}}
\newcommand\cnssymbol   {\ident{cons}}
\mathcommand\cnsindexsymbol[1]{\ident{cons}_{#1}}

\newcommand\lengthsymbol{\ident{length}}
\newcommand\dlsymbol    {\ident{dl}}
\newcommand\dloncesymbol{\ident{delonce}}
\newcommand\rcsymbol    {\ident{rc}}
\newcommand\brsymbol    {\ident{br}}
\newcommand\revtailsymbol{\ident{revtail}}
\newcommand\revsymbol{\ident{rev}}
\newcommand\appendsymbol {\ident{append}}
\newcommand\zeropredicatesymbol{\ident{zerop}}
\newcommand\eqsymbol        {\ident{eq}}
\newcommand\ifthensymbol    {\mbox{\ident{If{}Then}}}
\newcommand\ifthenelsesymbol{\mbox{\ident{If{}ThenElse}}}
\mathcommand\eqindexsymbol        [1]{\eqsymbol        _{#1}}
\mathcommand\ifthenindexsymbol    [1]{\ifthensymbol    _{#1}}
\mathcommand\ifthenelseindexsymbol[1]{\ifthenelsesymbol_{#1}}
\newcommand\orsymbol    {\ident{or}}
\newcommand\andsymbol   {\ident{and}}
\newcommand\leqsymbol   {\ident{leq}}
\newcommand\lessymbol   {\ident{less}}
\newcommand\lexsymbol   {\ident{lex}}
\newcommand\acksymbol   {\ident{ack}}
\newcommand\switchsymbol{\ident{switch}}
\newcommand\swatchsymbol{\ident{swatch}}
\newcommand\diveinssymbol{\ident{div1}}
\newcommand\divzweisymbol{\ident{div2}}
\newcommand\divrestsymbol{\ident{divrest}}
\newcommand\diveinstailsymbol{\ident{div1tail}}
\newcommand\divzweitailsymbol{\ident{div2tail}}

\newcommand\turingmachinesymbol{\ident T}
\newcommand\terminatespsymbol  {\ident{terminatesp}}
\newcommand\statesymbol        {\ident{state}}
\newcommand\cmdsymbol          {\ident{cmd}}
\newcommand\nthsymbol          {\ident{nth}}
\newcommand\doublesymbol       {\ident{double}}

\newcommand\ppsymbol           {\ident{p}}
\newcommand\qpsymbol           {\ident{q}}
\newcommand\Epsymbol           {\ident{E}}
\newcommand\Ppsymbol           {\ident{P}}
\newcommand\Qpsymbol           {\ident{Q}}
\newcommand\Marriessymbol      {\ident{Marries}}
\newcommand\Lovessymbol        {\ident{Loves}}
\newcommand\StolenBysymbol     {\ident{StolenBy}}
\newcommand\Humansymbol        {\ident{Human}}
\newcommand\Evensymbol         {\ident{Even}}
\newcommand\Oddsymbol          {\ident{Odd}}
\newcommand\Primesymbol        {\ident{Prime}}
\newcommand\EveryPairsymbol   {\ident{EveryPair}}
\newcommand\Givesymbol         {\ident{Give}}
\newcommand\Fathersymbol       {\ident{Father}}
\newcommand\Elephantpsymbol    {\ident{Elephant}}
\newcommand\Flowerpsymbol    {\ident{Flower}}
\newcommand\Germanpsymbol      {\ident{German}}
\newcommand\Bicyclepsymbol     {\ident{Bicycle}}
\newcommand\Hugepsymbol        {\ident{Huge}}
\newcommand\Animalpsymbol      {\ident{Animal}}
\newcommand\Malepsymbol        {\ident{Male}}
\newcommand\Boypsymbol        {\ident{Boy}}
\newcommand\Girlpsymbol        {\ident{Girl}}
\newcommand\Femalepsymbol      {\ident{Female}}
\newcommand\Roundpsymbol       {\ident{Round}}
\newcommand\Quadrangularpsymbol{\ident{Quadrangular}}
\newcommand\Metpsymbol         {\ident{Met}}
\newcommand\Bishopsymbol       {\ident{Bishop}}
\newcommand\mindexsymbol[1]{\existsvari w{#1}}

\newcommand\nonnegpsymbol      {\ident{nonnegp}}
\newcommand\wellsymbol         {\ident{well}}
\newcommand\welltailsymbol     {\ident{welltail}}
\newcommand\varsymbol          {\ident{var}}
\newcommand\aritysymbol        {\ident{arity}}

\newcommand\whilesymbol        {\ident{while}}

\newcommand\nullsymbol         {\ident{null}}
\newcommand\hdsymbol           {\ident{hd}}
\newcommand\tlsymbol           {\ident{tl}}
\newcommand\insymbol           {\ident{in}}
\newcommand\applysymbol        {\ident{app}}
\newcommand\termsymbol         {\ident{term}}
\mathcommand\tightim{\longrightarrow}
\mathcommand\im{\ \tightim\ }
\mathcommand\rs{\:\rulesugar\:\:}
\mathcommand\rulesugar{\longleftarrow}

\mathcommand\doublepp[1]      {\doublesymbol   \beginargs{#1}\allargs}
\mathcommand\aritypp[1]      {\aritysymbol   \beginargs{#1}\allargs}
\mathcommand\lengthpp[1]      {\lengthsymbol   \beginargs{#1}\allargs}
\mathcommand\wellpp[1]      {\wellsymbol   \beginargs{#1}\allargs}
\mathcommand\welltailpp[1]      {\welltailsymbol   \beginargs{#1}\allargs}
\mathcommand\varpp[1]      {\varsymbol   \beginargs{#1}\allargs}
\mathcommand\divrestpp[2]    {\divrestsymbol\beginargs{#1}\separgs{#2}\allargs}
\mathcommand\diveinspp[2]    {\diveinssymbol\beginargs{#1}\separgs{#2}\allargs}
\mathcommand\divzweipp[3]    {\divzweisymbol\beginargs{#1}\separgs{#2}
\separgs{#3}\allargs}
\mathcommand\diveinstailpp[4]    {\diveinstailsymbol\beginargs{#1}\separgs{#2}
\separgs{#3}\separgs{#4}\allargs}
\mathcommand\divzweitailpp[6]    {\divzweitailsymbol\beginargs{#1}\separgs{#2}
\separgs{#3}\separgs{#4}\separgs{#5}\separgs{#6}\allargs}
\mathcommand\mbppp[2]         {\mbpsymbol   \beginargs{#1}\separgs{#2}\allargs}
\mathcommand\revpp[1]     
{\revsymbol\beginargs{#1}\allargs}
\mathcommand\revppi[2]     
{\revsymbol^{#1}\beginargs{#2}\allargs}
\mathcommand\revtailpp[2]     
{\revtailsymbol\beginargs{#1}\separgs{#2}\allargs}
\mathcommand\revtailppi[3]
{\revtailsymbol^{#1}\beginargs{#2}\separgs{#3}\allargs}
\mathcommand\Permpp[2]     
{\Permsymbol\beginargs{#1}\separgs{#2}\allargs}
\mathcommand\Permppi[3]
{\Permsymbol^{#1}\beginargs{#2}\separgs{#3}\allargs}
\mathcommand\appendpp[2]      
{\appendsymbol \beginargs{#1}\separgs{#2}\allargs}
\mathcommand\appendppi[3]      
{\appendsymbol^{#1}\beginargs{#2}\separgs{#3}\allargs}
\mathcommand\Everypp[2]      
{\Everysymbol \beginargs{#1}\separgs{#2}\allargs}
\mathcommand\RExistspp[1]      
{\RExistssymbol \beginargs{#1}\allargs}
\mathcommand\appendlongpp[2]      
{\appendsymbol\left(\begin{array}{@{}l@{}}{#1}\separgs\\{#2}\end{array}\right)}
\mathcommand\cnspp[2]         {\cnssymbol   \beginargs{#1}\separgs{#2}\allargs}
\mathcommand\cnsppi[3]       {\cnssymbol^{#1}\beginargs{#2}\separgs{#3}\allargs}
\mathcommand\cnsindexpp[3]
{\cnsindexsymbol{#1}\beginargs{#2}\separgs{#3}\allargs}
\mathcommand\dlpp[2]          {\dlsymbol    \beginargs{#1}\separgs{#2}\allargs}
\mathcommand\dloncepp[2]      {\dloncesymbol\beginargs{#1}\separgs{#2}\allargs}
\mathcommand\dlonceppi[3]{\dloncesymbol^{#1}\beginargs{#2}\separgs{#3}\allargs}
\mathcommand\rcpp[2]          {\rcsymbol    \beginargs{#1}\separgs{#2}\allargs}
\mathcommand\brpp[2]          {\brsymbol    \beginargs{#1}\separgs{#2}\allargs}
\mathcommand\orpp[2]          {\orsymbol    \beginargs{#1}\separgs{#2}\allargs}
\mathcommand\andpp[2]         {\andsymbol   \beginargs{#1}\separgs{#2}\allargs}
\mathcommand\shortcnspp[2]    {\csymbol     \beginargs{#1}\separgs{#2}\allargs}
\mathcommand\tightshortcnspp[2]
{\csymbol\beginargs{#1}\tightsepargs{#2}\allargs}
\mathcommand\spp[1]           {\ssymbol     \beginargs{#1}\allargs}
\mathcommand\sppiterated[2]   {\ssymbol^{#1}\beginargs{#2}\allargs}
\mathcommand\ppp[1]           {\psymbol     \beginargs{#1}\allargs}
\mathcommand\pppiterated[2]   {\psymbol^{#1}\beginargs{#2}\allargs}
\mathcommand\zeropp           {\ident 0}
\mathcommand\Julietpp         {\ident{Juliet}}
\mathcommand\Romeopp          {\ident{Romeo}}
\mathcommand\Ipp              {\ident I}
\mathcommand\onepp            {\ident1}
\mathcommand\twopp            {\ident2}
\mathcommand\threepp          {\ident3}
\mathcommand\invertpp[1]      {\invertsymbol\beginargs{#1}\allargs}
\mathcommand\invpp[1]         {\invsymbol\beginargs{#1}\allargs}
\mathcommand\abspp[1]         {\abssymbol\beginargs{#1}\allargs}
\mathcommand\naturalspp[1]    {\naturalssymbol\beginargs{#1}\allargs}
\mathcommand\gensympp[1]      {\gensymsymbol\beginargs{#1}\allargs}
\mathcommand\nilpp            {\ident{nil}}
\mathcommand\falsepp          {\ident{false}}
\mathcommand\truepp           {\ident{true}}
\mathcommand\FALSEpp          {\ident{FALSE}}
\mathcommand\TRUEpp           {\ident{TRUE}}
\mathcommand\weirdppp         {\ident{weirdp}}
\mathcommand\ambigppp         {\ident{ambigp}}
\mathcommand\zeropredicatepp[1]{\zeropredicatesymbol\beginargs{#1}\allargs}
\mathcommand\cppeins       [1]{\csymbol     \beginargs{#1}\allargs}
\mathcommand\cppzwei       [2]{\csymbol\beginargs{#1}\separgs{#2}\allargs}
\mathcommand\eppeins       [1]{\esymbol     \beginargs{#1}\allargs}
\mathcommand\fppeins       [1]{\fsymbol     \beginargs{#1}\allargs}
\mathcommand\fppeinsindex  [2]{\fsymbol_{#1}\beginargs{#2}\allargs}
\mathcommand\fppeinsiterated[2]{\fsymbol^{#1}\beginargs{#2}\allargs}
\mathcommand\gppeins       [1]{\gsymbol     \beginargs{#1}\allargs}
\mathcommand\gppzwei       [2]{\gsymbol     \beginargs{#1}\separgs{#2}\allargs}
\mathcommand\hppeins       [1]{\hsymbol     \beginargs{#1}\allargs}
\mathcommand\kppeins       [1]{\ksymbol     \beginargs{#1}\allargs}
\mathcommand\appzero          {\ident a}
\mathcommand\bppzero          {\ident b}
\mathcommand\cppzero          {\ident c}
\mathcommand\dppzero          {\ident d}
\mathcommand\eppzero          {\ident e}
\mathcommand\eqindexpp[3]{\eqindexsymbol{#1}\beginargs{#2}\separgs{#3}\allargs}
\mathcommand\ifthenindexpp
[3]{\ifthenindexsymbol{#1}\beginargs{#2}\separgs{#3}\allargs}
\mathcommand\ifthenelseindexpp
[4]{\ifthenelseindexsymbol{#1}\beginargs{#2}\separgs{#3}\separgs{#4}\allargs}
\mathcommand\eqpp[2]{\eqsymbol\beginargs{#1}\separgs{#2}\allargs}
\mathcommand\leqpp[2]{\leqsymbol\beginargs{#1}\separgs{#2}\allargs}
\mathcommand\lespp[2]{\lessymbol\beginargs{#1}\separgs{#2}\allargs}
\mathcommand\lexpp[3]{\lexsymbol\beginargs{#1}\separgs{#2}\separgs{#3}\allargs}
\mathcommand\ackpp[2]{\acksymbol\beginargs{#1}\separgs{#2}\allargs}
\mathcommand\switchpp[1]{\switchsymbol\beginargs{#1}\allargs}
\mathcommand\swatchpp[1]{\swatchsymbol\beginargs{#1}\allargs}
\mathcommand\whilepp[2]{\whilesymbol\beginargs{#1}\separgs{#2}\allargs}
\mathcommand\nullpp[1]{\nullsymbol\beginargs{#1}\allargs}
\mathcommand\nullppiterated[2]{\nullsymbol^{#1}\beginargs{#2}\allargs}
\mathcommand\hdpp[1]{\hdsymbol\beginargs{#1}\allargs}
\mathcommand\hdppiterated[2]{\hdsymbol^{#1}\beginargs{#2}\allargs}
\mathcommand\tlpp[1]{\tlsymbol\beginargs{#1}\allargs}
\mathcommand\tlppiterated[2]{\tlsymbol^{#1}\beginargs{#2}\allargs}
\mathcommand\inpp[2]{\insymbol\beginargs{#1}\separgs{#2}\allargs}
\mathcommand\inppiterated[3]{\insymbol^{#1}\beginargs{#2}\separgs{#3}\allargs}
\mathcommand\applypp[2]{\applysymbol\beginargs{#1}\separgs{#2}\allargs}
\mathcommand\applyppiterated
[3]{\applysymbol^{#1}\beginargs{#2}\separgs{#3}\allargs}
\mathcommand\termpp[2]{\termsymbol\beginargs{#1}\separgs{#2}\allargs}
\mathcommand\setpp[1]{\set\beginargs{#1}\allargs}

\mathcommand\Tpp[6]{\turingmachinesymbol\beginargs{#1}\separgs{#2}\separgs
{#3}\separgs{#4}\separgs{#5}\separgs{#6}\allargs}
\mathcommand\Tppseven[7]{\turingmachinesymbol\beginargs{#1}\separgs{#2}\separgs
{#3}\separgs{#4}\separgs{#5}\separgs{#6}\separgs{#7}\allargs}
\mathcommand\foreverppp[6]{\ident{foreverp}\beginargs{#1}\separgs{#2}\separgs
{#3}\separgs{#4}\separgs{#5}\separgs{#6}\allargs}
\mathcommand\terminatesppp[6]{\terminatespsymbol\beginargs{#1}\separgs
{#2}\separgs{#3}\separgs{#4}\separgs{#5}\separgs{#6}\allargs}
\mathcommand\terminatespppone[1]{\terminatespsymbol \beginargs{#1}\allargs}
\mathcommand\statepp
[3]{\statesymbol\beginargs{#1}\separgs{#2}\separgs{#3}\allargs}
\mathcommand\tightstatepp
[3]{\statesymbol\beginargs{#1}\tightsepargs{#2}\tightsepargs{#3}\allargs}
\mathcommand\cmdpp
[3]{\cmdsymbol  \beginargs{#1}\separgs{#2}\separgs{#3}\allargs}
\mathcommand\tightcmdpp
[3]{\cmdsymbol  \beginargs{#1}\tightsepargs{#2}\tightsepargs{#3}\allargs}
\mathcommand\stoppp           {\ident{stop}}
\mathcommand\leftpp           {\ident{left}}
\mathcommand\rightpp          {\ident{right}}
\mathcommand\nthpp         [2]{\nthsymbol  \beginargs{#1}\separgs{#2}\allargs}
\mathcommand\pppp          [1]{\ppsymbol\beginargs{#1}            \allargs}
\mathcommand\qppp          [2]{\qpsymbol\beginargs{#1}\separgs{#2}\allargs}
\mathcommand\Eppp          [1]{\Epsymbol\beginargs{#1}            \allargs}
\mathcommand\Epppzwei      [2]{\Epsymbol\beginargs{#1}\separgs{#2}\allargs}
\mathcommand\Pppp          [1]{\Ppsymbol\beginargs{#1}            \allargs}
\mathcommand\Ppppdrei      
[3]{\Ppsymbol\beginargs{#1}\separgs{#2}\separgs{#3}\allargs}
\mathcommand\Ppppvier
[4]{\Ppsymbol\beginargs{#1}\separgs{#2}\separgs{#3}\separgs{#4}\allargs}
\mathcommand\Qppp          [2]{\Qpsymbol\beginargs{#1}\separgs{#2}\allargs}
\mathcommand\Qpppeins      [1]{\Qpsymbol\beginargs{#1}\allargs}
\mathcommand\Qpppdrei      
[3]{\Qpsymbol\beginargs{#1}\separgs{#2}\separgs{#3}\allargs}
\mathcommand\Fatherpp      [2]{\Fathersymbol\beginargs{#1}\separgs{#2}\allargs}
\mathcommand\Marriespp     [2]{\Marriessymbol\beginargs{#1}\separgs{#2}\allargs}
\mathcommand\Lovespp       [2]{\Lovessymbol\beginargs{#1}\separgs{#2}\allargs}
\mathcommand\StolenBypp    [2]
{\StolenBysymbol\beginargs{#1}\separgs{#2}\allargs}
\mathcommand\Humanpp       [1]{\Humansymbol\beginargs{#1}\allargs}
\mathcommand\Evenpp        [1]{\Evensymbol\beginargs{#1}\allargs}
\mathcommand\Evenppi       [2]{\Evensymbol^{#1}\beginargs{#2}\allargs}
\mathcommand\Oddpp         [1]{\Oddsymbol\beginargs{#1}\allargs}
\mathcommand\Primepp       [1]{\Primesymbol\beginargs{#1}\allargs}
\mathcommand\EveryPairpp  [2]{\EveryPairsymbol\beginargs{#1}\separgs
{#2}\allargs}
\mathcommand\mindexppeins  [2]{\mindexsymbol{#1}\beginargs{#2}\allargs}
\mathcommand\Givepp        [3]{\Givesymbol
\beginargs{#1}\separgs{#2}\separgs{#3}\allargs}
\mathcommand\mindexppzwei  [3]{\mindexsymbol
{#1}\beginargs{#2}\separgs{#3}\allargs}
\mathcommand\mindexppdrei  [4]{\mindexsymbol
{#1}\beginargs{#2}\separgs{#3}\separgs{#4}\allargs}

\mathcommand\nonnegppp     [1]{\nonnegpsymbol\beginargs{#1}\allargs}

\mathcommand\anonymouscsymbol{c}
\mathcommand\anonymouscindexsymbol[1]{\anonymouscsymbol_{#1}}
\mathcommand\anonymousfsymbol{f}
\mathcommand\anonymouscpp
[2]{\anonymouscsymbol\beginargs{#1}\separgs\ldots\separgs{#2}\allargs}
\mathcommand\anonymouscindexpp
[3]{\anonymouscindexsymbol{#1}\beginargs{#2}\separgs\ldots\separgs{#3}\allargs}
\mathcommand\anonymousfpp
[2]{\anonymousfsymbol\beginargs{#1}\separgs\ldots\separgs{#2}\allargs}
\mathcommand\coerceindexpp[3]{[#3]_{#1}^{#2}}

\mathcommand\Elephantppp    [1]{\Elephantpsymbol\beginargs{#1}\allargs}
\mathcommand\Flowerppp      [1]{\Flowerpsymbol  \beginargs{#1}\allargs}
\mathcommand\Bicycleppp     [1]{\Bicyclepsymbol \beginargs{#1}\allargs}
\mathcommand\Germanppp      [1]{\Germanpsymbol  \beginargs{#1}\allargs}
\mathcommand\Hugeppp        [1]{\Hugepsymbol    \beginargs{#1}\allargs}
\mathcommand\Animalppp      [1]{\Animalpsymbol  \beginargs{#1}\allargs}
\mathcommand\Maleppp        [1]{\Malepsymbol    \beginargs{#1}\allargs}
\mathcommand\Boyppp         [1]{\Boypsymbol     \beginargs{#1}\allargs}
\mathcommand\Girlppp        [1]{\Girlpsymbol    \beginargs{#1}\allargs}
\mathcommand\Femaleppp      [1]{\Femalepsymbol  \beginargs{#1}\allargs}
\mathcommand\Roundppp       [1]{\Roundpsymbol   \beginargs{#1}\allargs}
\mathcommand\Bishoppp       [1]{\Bishopsymbol   \beginargs{#1}\allargs}
\mathcommand\Quadrangularppp[1]{\Quadrangularpsymbol  \beginargs{#1}\allargs}
\mathcommand\Metppp[2]{\Metpsymbol     \beginargs{#1}\separgs{#2}\allargs}

\newcommand\bound     {{\rm bound}}
\newcommand\free      {{\rm free}}

\mathcommand\Vtripleindex[3]{\V\!_{{#1},\,{#2},\,{#3}}}
\mathcommand\Vdoubleindex[2]{\V\!_{{#1},\,{#2}}}
\mathcommand\Vsingleindex[1]{\V\!_{{#1}}}

\mathcommand\Erel[1]{\Gammaoffont\!_{#1}}
\mathcommand\Urel[1]{\Deltaoffont_{#1}}

\newcommand\Rsub    {\math R-sub\-sti\-tu\-tion}

\newcommand\RSub    {\math R-Sub\-sti\-tu\-tion}

\newcommand\strongexvalof[2]{\pair{#1}{#2}-valu\-a\-tion}

\newcommand\strongexRval {\strongexvalof\salgebra R}

\newcommand\cc{choice-condition}

\newcommand\vc{vari\-able-con\-di\-tion}
\newcommand\VC{Vari\-able-Con\-di\-tion}
\newcommand\Vc{Vari\-able-con\-di\-tion}

\mathcommand\theRprimefromstrongtoweak{
  \inparenthesesinlinetight{
     \domres\id{\Vwall\cup\Vsome\setminus\RAN\varsigma}
     \nottight{\nottight\uplus}
     \reverserelation\varsigma
  }
  \nottight{\circ}
  \ranres
    {\transclosureinline R}
    {\Vwall\cup\Vsome\setminus\RAN\varsigma}
  \nottight{\nottight{\nottight{\uplus}}}
  \Vsome\tighttimes\Vsall
}

\mathcommand\deltaminus{\delta^-}
\mathcommand\deltaplus{\delta^+}
\mathcommand\deltaplusplus{\delta^{+^+}}
\mathcommand\deltastar{\delta^*}
\mathcommand\deltastarstar{\delta^{*^*}}
\newcommand\varihelper[1]{free \discretionary
{\mbox{\math{#1}-vari-}}{\mbox{able}}{\mbox{\math{#1}-variable}}}

\newcommand\fev {\varihelper\gamma}
\newcommand\fuv {\varihelper\delta}
\newcommand\wfuv{\varihelper{\delta^-}}
\newcommand\sfuv{\varihelper{\delta^+}}

\newcommand\SFUV {Free \math{\delta^+}-Variable}

\mathcommand\Vall     {\Vsingleindex\indexdelta         }
\mathcommand\Vwall    {\Vsingleindex\indexdeltaminu     }
\mathcommand\Vsall    {\Vsingleindex\indexdeltaplus     }
\mathcommand\Vgsome   {\Vsingleindex\indexgammaplus     }
\mathcommand\Vsome    {\Vsingleindex\indexgamma         }
\mathcommand\Vfree    {\Vsingleindex\indexfree          }
\mathcommand\Vbound   {\Vsingleindex\indexbound         }
\mathcommand\Vsomesall{\Vsingleindex\indexgammadeltaplus}

\mathapplycommand\VARall      {\VARsingleindex\indexdelta         }
\mathapplycommand\VARwall     {\VARsingleindex\indexdeltaminu     }
\mathapplycommand\VARsall     {\VARsingleindex\indexdeltaplus     }
\mathapplycommand\VARgsome    {\VARsingleindex\indexgammaplus     }
\mathapplycommand\VARsome     {\VARsingleindex\indexgamma         }
\mathapplycommand\VARfree     {\VARsingleindex\indexfree          }
\mathapplycommand\VARbound    {\VARsingleindex\indexbound         }
\mathapplycommand\VARsomesall {\VARsingleindex\indexgammadeltaplus}
\mathcommand\displayVARsall[1]{\VARsingleindex\indexdeltaplus
\!\!\!\:\left(\begin{array}{@{}c@{}}#1\end{array}\right)}

\mathcommand\rigidvari     [2]{#1_{#2}^\indexgammadeltaplus}
\mathcommand\existsvari    [2]{#1_{#2}^\indexgamma    }
\mathcommand\forallvari    [2]{#1_{#2}^\indexdelta    }
\mathcommand\freevari      [2]{#1_{#2}^\indexfree     }
\mathcommand\wforallvari   [2]{#1_{#2}^\indexdeltaminu}
\mathcommand\sforallvari   [2]{#1_{#2}^\indexdeltaplus}
\mathcommand\gexistsvari   [2]{#1_{#2}^\indexgammaplus}
\mathcommand\boundvari     [2]{#1_{#2}}
\mathcommand\vari          [2]{#1_{#2}}
\mathcommand\wforallvarilow[2]{#1_{#2}^
{\raisebox{-.82ex}{\math\indexdeltaminu}}}

\newcommand\indexhelper[1]{{\scriptscriptstyle#1\:\!\!}}
\newcommand\indexdeltaplus
{\indexhelper{\delta^{\raisebox{-.17ex}{\fvesf\hskip-0.14em +}}}}
\newcommand\indexdeltaminu
{\indexhelper{\delta^{\mbox{\fvesf\hskip-0.14em\rule[.2ex]{.7em}{.15ex}}}}}
\newcommand\indexgammaplus
{\indexhelper{\gamma^{\mbox{\fvesf\hskip-0.14em +}}}}
\newcommand\indexgammadeltaplus
{\indexhelper{\gamma\delta^{\raisebox{-.17ex}{\fvesf\hskip-0.14em +}}}}

\newcommand\indexdelta{\indexhelper\delta}
\newcommand\indexgamma{\indexhelper\gamma}
\newcommand\indexfree
{{\scriptscriptstyle\free}}
\newcommand\indexbound
{{\scriptscriptstyle\bound}}

\newcommand\Wellfsymb{\ident{Wellf}}
\mathapplycommand\Wellfpp{\Wellfsymb}
\mathcommand\beginargs{(}
\mathcommand\allargs  {)}
\mathcommand\separgs  {,\,}
\mathcommand\tightsepargs{,}

\mathcommand\minusppnoparentheses  [2]{{#1}\,\minussymbol\,{#2}}
\mathcommand\tightminusppnoparentheses  [2]{{#1}\minussymbol{#2}}
\mathcommand\divideppnoparentheses [2]{{#1}\,\dividesymbol\,{#2}}
\mathcommand\plusppnoparentheses   [2]{{#1}\,\plussymbol \,{#2}}
\mathcommand\plusppnoparenthesesi  [3]{{#2}\,\plussymbol^{#1}\,{#3}}
\mathcommand\tightplusppnoparentheses   [2]{{#1}\plussymbol{#2}}
\mathcommand\timesppnoparentheses  [2]{{#1}\,\timessymbol\,{#2}}
\mathcommand\undppnoparentheses    [2]{{#1}\und            {#2}}
\mathcommand\oderppnoparentheses   [2]{{#1}\oder           {#2}}
\mathcommand\impliesppnoparentheses[2]{{#1}\implies        {#2}}
\mathcommand\leqinfixppnoparentheses[2]{{#1}\,\tight\leq\,{#2}}
\mathcommand\geqinfixppnoparentheses[2]{{#1}\,\tight\geq\,{#2}}
\mathcommand\dividepp [2]{(\divideppnoparentheses {#1}{#2})}
\mathcommand\minuspp  [2]{(\minusppnoparentheses  {#1}{#2})}
\mathcommand\pluspp   [2]{(\plusppnoparentheses   {#1}{#2})}
\mathcommand\timespp  [2]{(\timesppnoparentheses  {#1}{#2})}
\mathcommand\undpp    [2]{(\undppnoparentheses    {#1}{#2})}
\mathcommand\oderpp   [2]{(\oderppnoparentheses   {#1}{#2})}
\mathcommand\impliespp[2]{(\impliesppnoparentheses{#1}{#2})}

\catcode`\@=11
\input ENDNOTES.sty
\catcode`\@=12
\def\enotesize{\normalsize}
\def\endnoteendsep{\vspace{5.0\topsep}}
\let\footnote=\endnote
\input xy
\xyoption{ps}
\xyoption{all}
\xyoption{dvips}
\usepackage{url}

\newlength{\mybibitemsep}
\newcommand\setmybibitemsep[1]{\setlength{\mybibitemsep}{#1}}

\setmybibitemsep{0.5\topsep}

\newcommand\includenetreferences{y}

\newcommand\referencessize{\normalsize}

\newcommand\mybibbaselinestretch{0.96}

\newcommand\mybibsection[1]{\section*{#1}\if 
\addcontentslineofbibsection\addcontentsline{toc}{section}{#1}\fi}
\newcommand\addcontentslineofbibsection{y}

\newcommand\mybibtitle[1]{{\em #1\@.}}
\newcommand\mybibhardnodate
[1]{{\def\myfirstarg{#1}\ifx\empty\myfirstarg{}\else#1.\ \fi}}
\newcommand\mybibsoft
[2]{{\def\myfirstarg{#1}\def\mysecondarg{#2}\ifx\empty\myfirstarg{}\else
\if y\includenetreferences\writemynetsource#1\else\if x\includenetreferences
\ifx\empty\mysecondarg\writemynetsource#1\else{}\fi\else
\if n\includenetreferences{}\else\typeout
{WARNING includenetreferences from headerbiblio.tex has silly definition}
\fi\fi\fi\fi}}

\def\writemynetsource[#1,#2,#3,#4://#5]{{\tt\sloppy\ 
\url{#4://#5} \discretionary
{(\ignorespaces#2\,\ignorespaces#1\mbox{$\!$},\mbox{$\!$}}%
{\ignorespaces#3)\mbox{$\!$}.}%
{(\ignorespaces#2\,\ignorespaces#1\mbox{$\!$},\,\ignorespaces
#3)\mbox{$\!$}.}}}

\catcode`\@=11

\newcommand\resetlongbibstyle{%
\def\bibitem{\@ifnextchar[\@lbibitem\@bibitem}
\def\@lbibitem[##1]##2{\item[\@biblabel{##1}\hfill]\if@filesw
      {\let\protect\noexpand
       \immediate
       \write\@auxout{\string\bibcite{##2}{##1}}}\fi\ignorespaces}
\def\@bibitem##1{\item\if@filesw \immediate\write\@auxout
       {\string\bibcite{##1}{\the\value{\@listctr}}}\fi\ignorespaces}
\def\bibcite{\@newl@bel b}
\let\citation\@gobble
\let\bibdata=\@gobble
\let\bibstyle=\@gobble
\def\bibliography##1{%
  \if@filesw
    \immediate\write\@auxout{\string\bibdata{##1}}%
  \fi
  \@input@{\jobname.bbl}}
\def\bibliographystyle##1{%
  \ifx\@begindocumenthook\@undefined\else
    \expandafter\AtBeginDocument
  \fi
    {\if@filesw
       \immediate\write\@auxout{\string\bibstyle{##1}}%
     \fi}}
\def\nocite##1{\@bsphack
  \@for\@citeb:=##1\do{%
    \edef\@citeb{\expandafter\@firstofone\@citeb}%
    \if@filesw\immediate\write\@auxout{\string\citation{\@citeb}}\fi
    \@ifundefined{b@\@citeb}{\G@refundefinedtrue
        \@latex@warning{Citation `\@citeb' undefined}}{}}%
  \@esphack}
\expandafter\let\csname b@*\endcsname\@empty
\DeclareRobustCommand\cite{%
  \@ifnextchar [{\@tempswatrue\@citex}{\@tempswafalse\@citex[]}}
\DeclareRobustCommand\citet{%
  \@ifnextchar [{\@tempswatrue\@citex}{\@tempswafalse\@citex[]}}
\def\@tempswafalse{\let\if@tempswa\iffalse}
\def\@tempswatrue{\let\if@tempswa\iftrue}
\let\if@tempswa\iffalse
\def\@cite##1##2{##1\if@tempswa , ##2\fi}
\def\@citex[##1]##2{%
  \let\@citea\@empty
  \@cite{\@for\@citeb:=##2\do
    {\@citea\def\@citea{,\penalty\@m\ }%
     \edef\@citeb{\expandafter\@firstofone\@citeb}%
     \if@filesw\immediate\write\@auxout{\string\citation{\@citeb}}\fi
     \@ifundefined{b@\@citeb}{\mbox{\reset@font\bfseries ?}%
       \G@refundefinedtrue
       \@latex@warning
         {Citation `\@citeb' on page \thepage \space undefined}}%
       {\csname b@\@citeb\endcsname}}}{##1}}
\def\@biblabel##1{##1}
\def\mybibitem##1##2##3##4##5##6##7##8##9
{\item
 \if@filesw
      {\let\protect\noexpand
       \immediate
       \write\@auxout{\string\bibcite{##1}{##7\discretionary
{}{}{\,}(##3##9)}}}\fi\ignorespaces
##2 (##3##9). \mybibtitle{##4} \mybibhardnodate{##5}\mybibsoft
{##6}{##5}$\!\!$\par}
\def\thebibliography##1{\mybibsection{\refname}%
\@mkboth{\uppercase{\refname}}{\uppercase{\refname}}%\vskip\referencesfootroom
\def\baselinestretch{\mybibbaselinestretch}%
\list{}{\labelwidth\z@
    \leftmargin 1.5pc
    \itemindent-\leftmargin}
    \referencessize
    \parindent\z@
    \parskip\mybibitemsep\relax
    \def\newblock{\hskip .11em plus .33em minus .07em}
    \sloppy\clubpenalty4000\widowpenalty4000
    \sfcode`\.=1000\relax}
\let\endthebibliography=\endlist
}

\newcommand\resetshortbibstyle{%
\def\bibitem{\@ifnextchar[\@lbibitem\@bibitem}
\def\@lbibitem[##1]##2{\item[\@biblabel{##1}\hfill]\if@filesw
      {\let\protect\noexpand
       \immediate
       \write\@auxout{\string\bibcite{##2}{##1}}}\fi\ignorespaces}
\def\@bibitem##1{\item\if@filesw \immediate\write\@auxout
       {\string\bibcite{##1}{\the\value{\@listctr}}}\fi\ignorespaces}
\def\bibcite{\@newl@bel b}
\let\citation\@gobble
\def\@citex[##1]##2{%
  \let\@citea\@empty
  \@cite{\@for\@citeb:=##2\do
    {\@citea\def\@citea{,\penalty\@m\ }%
     \edef\@citeb{\expandafter\@firstofone\@citeb}%
     \if@filesw\immediate\write\@auxout{\string\citation{\@citeb}}\fi
     \@ifundefined{b@\@citeb}{\mbox{\reset@font\bfseries ?}%
       \G@refundefinedtrue
       \@latex@warning
         {Citation `\@citeb' on page \thepage \space undefined}}%
       {\hbox{\csname b@\@citeb\endcsname}}}}{##1}}
\let\bibdata=\@gobble
\let\bibstyle=\@gobble
\def\bibliography##1{%
  \if@filesw
    \immediate\write\@auxout{\string\bibdata{##1}}%
  \fi
  \@input@{\jobname.bbl}}
\def\bibliographystyle##1{%
  \ifx\@begindocumenthook\@undefined\else
    \expandafter\AtBeginDocument
  \fi
    {\if@filesw
       \immediate\write\@auxout{\string\bibstyle{##1}}%
     \fi}}
\def\nocite##1{\@bsphack
  \@for\@citeb:=##1\do{%
    \edef\@citeb{\expandafter\@firstofone\@citeb}%
    \if@filesw\immediate\write\@auxout{\string\citation{\@citeb}}\fi
    \@ifundefined{b@\@citeb}{\G@refundefinedtrue
        \@latex@warning{Citation `\@citeb' undefined}}{}}%
  \@esphack}
\expandafter\let\csname b@*\endcsname\@empty
\def\@cite##1##2{[{##1\if@tempswa , ##2\fi}]}
\def\@biblabel##1{[##1]}
\DeclareRobustCommand\cite{%
  \@ifnextchar [{\@tempswatrue\@citex}{\@tempswafalse\@citex[]}}
\def\@tempswafalse{\let\if@tempswa\iffalse}
\def\@tempswatrue{\let\if@tempswa\iftrue}
\let\if@tempswa\iffalse
\def\mybibitem##1##2##3##4##5##6##7##8##9
{\bibitem[##8##9]{##1}##2 (##3). 
\mybibtitle{##4} \mybibhardnodate{##5}\mybibsoft{##6}{##5}$\!\!$\par}
\def\thebibliography##1{\mybibsection{\refname}%
\@mkboth{\uppercase{\refname}}{\uppercase{\refname}}%
\def\baselinestretch{\mybibbaselinestretch}%
\referencessize
\list
 {[\arabic{enumi}]}
 {\settowidth\labelwidth{[##1]}\leftmargin\labelwidth
 \advance\leftmargin\labelsep
 \usecounter{enumi}}
 \parskip\mybibitemsep\relax
 \def\newblock{\hskip .11em plus .33em minus .07em}
 \sloppy\clubpenalty4000\widowpenalty4000
 \sfcode`\.=1000\relax}
\let\endthebibliography=\endlist
}

\newcommand\resetnumberbibstyle{%
\def\bibitem{\@ifnextchar[\@lbibitem\@bibitem}
\def\@lbibitem[##1]##2{\item[\@biblabel{##1}\hfill]\if@filesw
      {\let\protect\noexpand
       \immediate
       \write\@auxout{\string\bibcite{##2}{##1}}}\fi\ignorespaces}
\def\@bibitem##1{\item\if@filesw \immediate\write\@auxout
       {\string\bibcite{##1}{\the\value{\@listctr}}}\fi\ignorespaces}
\def\bibcite{\@newl@bel b}
\let\citation\@gobble
\def\@citex[##1]##2{%
  \let\@citea\@empty
  \@cite{\@for\@citeb:=##2\do
    {\@citea\def\@citea{,\penalty\@m\ }%
     \edef\@citeb{\expandafter\@firstofone\@citeb}%
     \if@filesw\immediate\write\@auxout{\string\citation{\@citeb}}\fi
     \@ifundefined{b@\@citeb}{\mbox{\reset@font\bfseries ?}%
       \G@refundefinedtrue
       \@latex@warning
         {Citation `\@citeb' on page \thepage \space undefined}}%
       {\hbox{\csname b@\@citeb\endcsname}}}}{##1}}
\let\bibdata=\@gobble
\let\bibstyle=\@gobble
\def\bibliography##1{%
  \if@filesw
    \immediate\write\@auxout{\string\bibdata{##1}}%
  \fi
  \@input@{\jobname.bbl}}
\def\bibliographystyle##1{%
  \ifx\@begindocumenthook\@undefined\else
    \expandafter\AtBeginDocument
  \fi
    {\if@filesw
       \immediate\write\@auxout{\string\bibstyle{##1}}%
     \fi}}
\def\nocite##1{\@bsphack
  \@for\@citeb:=##1\do{%
    \edef\@citeb{\expandafter\@firstofone\@citeb}%
    \if@filesw\immediate\write\@auxout{\string\citation{\@citeb}}\fi
    \@ifundefined{b@\@citeb}{\G@refundefinedtrue
        \@latex@warning{Citation `\@citeb' undefined}}{}}%
  \@esphack}
\expandafter\let\csname b@*\endcsname\@empty
\def\@cite##1##2{[{##1\if@tempswa , ##2\fi}]}
\def\@biblabel##1{[##1]}
\DeclareRobustCommand\cite{%
  \@ifnextchar [{\@tempswatrue\@citex}{\@tempswafalse\@citex[]}}
\def\@tempswafalse{\let\if@tempswa\iffalse}
\def\@tempswatrue{\let\if@tempswa\iftrue}
\let\if@tempswa\iffalse
\def\mybibitem##1##2##3##4##5##6##7##8##9
{\bibitem{##1}##2 (##3). 
\mybibtitle{##4} \mybibhardnodate{##5}\mybibsoft{##6}{##5}$\!\!$\par}
\def\thebibliography##1{\mybibsection{\refname}%
\@mkboth{\uppercase{\refname}}{\uppercase{\refname}}
\def\baselinestretch{\mybibbaselinestretch}%
\referencessize
\list
 {[\arabic{enumi}]}
 {\settowidth\labelwidth{[##1]}\leftmargin\labelwidth
 \advance\leftmargin\labelsep
 \usecounter{enumi}}
 \parskip\mybibitemsep\relax
 \def\newblock{\hskip .11em plus .33em minus .07em}
 \sloppy\clubpenalty4000\widowpenalty4000
 \sfcode`\.=1000\relax}
\let\endthebibliography=\endlist
}

\newcommand\setnumberbibstyle{\AtBeginDocument{\resetnumberbibstyle}}

\catcode`\@=12

\setnumberbibstyle
\renewcommand\mybibbaselinestretch{1.0}
\setmybibitemsep{1.0\topsep}
\renewcommand\mybibsection[1]{\mysection #1\yestop}
\renewcommand\referencessize{\normalsize}
\date
{{\small First Print Edition: August\,8, 2005}
\\{\small Thoroughly Updated \Feb\,27, 2006}
\\{\small Minorly Improved \Jul\,30, 2006}\SEKIedition}
\catcode`\@=11
\renewcommand\tableofcontents{%
    \section*{\contentsname
        \@mkboth{%
           \MakeUppercase\contentsname}{\MakeUppercase\contentsname}}%
    \halftop
    \@starttoc{toc}%
    }
\def\l@section#1#2{%
  \ifnum \c@tocdepth >\z@
    \addpenalty\@secpenalty
    \addvspace{1.0em \@plus\p@}%
    \setlength\@tempdima{2.5em}%
    \begingroup
      \parskip -8pt
      \parindent \z@ \rightskip \@pnumwidth
      \parfillskip -\@pnumwidth
      \leavevmode
      \advance\leftskip\@tempdima
      \hskip -\leftskip
      #1\nobreak\hfil \nobreak\hb@xt@\@pnumwidth{\hss #2}\par
    \endgroup
  \fi}
\catcode`\@=12
\begin{document}
\setcounter{tocdepth}{1}\makecover
% Let your text start here, possibly changing the following a lot.
\maketitle
\begin{abstract}%
Using a human-oriented formal example proof of the (\math{\lim\tight+}) theorem,
\ie\ that \whatproofisabout,
which is of value for reference on its own,
we exhibit a \nonpermutability\ of \math\beta-steps and \mbox{\deltaplus-steps}
(according to \smullyan's classification), which is not visible with
non-liberalized \math\delta-rules
and not serious with further liberalized \mbox{\math\delta-rules},
such as the \deltaplusplus-rule.
Besides a careful presentation of the search for a proof of
\nlbmath{(\lim\tight+)} with several pedagogical intentions,
the main subject is to explain why the order of
\math\beta-steps plays such a practically important role in some calculi.\end
{abstract}\yestop\halftop\tableofcontents\vfill\pagebreak

\section{Motivation}\label{section motivation}

\begin{sloppypar}
In December\,2004, 
in the theoretical part of an advanced senior-level 
lecture course \nlbcite{maslecture} on \maslong s, \ I \nolinebreak presented
a formal example proof in a human-oriented sequent calculus
that \whatproofisabout\ (\math{\lim\tight+}). \ 
{\em\Maslong s}
are human-oriented interactive theorem provers with strong automation support,
aiming at a synergetic interplay between mathematician and machine. \ 
\PVS\ \nlbcite{pvs},
\OMEGA\ \nlbcite{omegacadetwo},
\ISABELLEHOL\ \nlbcite{isabellehol,isabellesevenhundred},
and \QUODLIBET\ \nlbcite{quodlibet-cade}
are some of the systems approaching this long term goal.
%Although a previous version of the (\math{\lim\tight+}) proof was already part 
%of a lecture course I gave a year before \nlbcite{wirthlecture},
%It was again quite some heavy \LaTeX\ work to prepare the slides for the
%lecture. You find the complete proof now in \sectref{section proof}.
\end{sloppypar}
Considering\emph{reductive calculi} such as sequent, tableau, or matrix calculi,
one of the functions of my lectures within the course was to show 
that---although\emph{sequents} are easier to understand due to their 
locality---\tightemph{matrixes} 
(or\emph{indexed formula trees} \nlbcite{sergediss,wallen})
are not only a clever implementation,
but---more importantly for us---also needed to follow
the proof organization of a working mathematician. \ 
%; whereas the\emph{tableau} version lies somewhere in the middle.
To this end, \
I \nolinebreak tried to give the students an idea of the 
premature commitments forced by sequent and tableau calculi,
which require a mathematician 
to deviate from his intended proof plans and proof-search heuristics.
%intentional planning and organization of the proof's search space.

In his fascinating book \nlbcite{wallen}, \
\wallenname\ had criticized 
the \nonpermutability\ of \math\gamma- and \math\delta-steps
in sequent calculi,
according to \smullyanname's classification 
and uniform notation of reductive inference rules as
\math\alpha, \nlbmath\beta, \math\gamma, and \nlbmath\delta\ \cite{smullyan}. \ 
I explained how this \nonpermutability\ can be overcome 
by replacing the (non-liberalized) \math{\delta}-rule 
(which we will call\emph{\deltaminus-rule})
with the liberalized \deltaplus-rule \nlbcite{deltaplus}. \ 
Along the \math{(\lim\tight+)} proof, \ 
I then showed that with the \deltaplus-rule, however, another 
\nonpermutability\ becomes visible,
now of the \mbox{\math\beta- and \deltaplus-steps}. \
Before the liberalization took place to make logicians glad,\emph
{this \nonpermutability\ was hidden behind the  \nonpermutability\ of the 
 \math\gamma- \nolinebreak and \nlbmath\deltaminus-steps.}\footnote
{\label{note scornful}A scornful anonymous 
 referee of a previous version of this 
 \daspaper\
 (who was the only one to reject it for the \thefourteenthTABLEAUfive) 
  wrote:\begin{quote}``For once a positive comment: 
   The first lines of page\,12 finally
   contain a very interesting insight, namely that different
   \nonpermutabilities\ can
   hide each other.''\end{quote}}

At \nolinebreak that moment, the best logician among my co-lecturers 
contradicted the occurrence of this \nonpermutability,
and insisted on his opinion 
when I \nolinebreak repeated 
the material for an introduction in the next lecture.
Thus, the \nonpermutability\ problems of \math\beta-steps deserve publication.
A \nolinebreak referee of a previous version of this \daspaper\ called this
``an interesting but not too surprising result\closequotefullstop
Besides this hard result, following the lecture,
in this \daspaper\ we will address some soft aspects of formal
calculi for human--machine interaction and publish
(for the first time?)\ 
a more or less readable, complete, and human-oriented
proof of a mathematical standard theorem
in a standard general-purpose formal calculus in \sectref{section proof}. \ 
We \nolinebreak discuss the \nonpermutabilities\ of this example proof
in \sectref{section discussion}, 
prove the \nonpermutability\ of its crucial \math\beta- and \deltaplus-step
in \sectref{section meta-proof}, and conclude with an emphasis on open problems
in \sectref{section conclusion}.

\knuthquotation{\fraknomath Zuerst werden die Leute eine Sache leugnen;
dann werden sie sie verharmlosen;\\dann werden sie beschlie\sz en,
sie sei seit langem bekannt.}{\humboldtname\ \rm(cited according to \cite{muehle-des-hamlet}, \p\,x)}
\vfill\cleardoublepage
%%%%%%%%%%%%%%%%%%%%%%%%%%%%%%%%%%%%%%%%%%%%%%%%%%%%%%%%%%%%%%%%%%%%%%%%%%%%%%%
\section
{Introduction to \NonPermutabilities\ \etc}\label{section introduction}
As explained in \cite{wallen},
the search space of sequent or tableau calculi
may suffer from the following weaknesses in design:
\emph{Irrelevance},\emph{Notational Redundancy}, and\emph\NonPermutability.
Unless explicitly stated otherwise,
the weaknesses described in the following apply to sequent and tableau
calculi alike.

\yestop\noindent{\bf Irrelevance} means, \eg, that when proving the sequent
\par\noindent\LINEmath{A\comma\ 
  \neg\inpit{B\und\Lovespp\Romeopp{\existsvari y 0}}\comma\ 
  \Lovespp\Romeopp\Julietpp}\par\noindent
with \math A and \math B some big formulas,
we may try to prove \math A or \math{\neg B} 
for a long time, \linebreak although this is not relevant
if they are false.
Note that in this \daspaper\emph{sequents} are just lists of formulas,
\ie\ the simplest form that will do for two-valued logics.
We call\emph{\fev s} 
(after the \math\gamma-steps, which may introduce new ones)
(written as \nlbmath{\existsvari y 0}) what has the standard names
of ``meta'' \nlbcite{isabellehol} 
or ``free'' \nlbcite{fitting} variables.
Indeed, \fev s must be distinguished
from the true meta-variables and 
the other kinds of free variables we will need.
The means to avoid irrelevance is focusing on\emph{connections},
just as the one between 
\bigmaths{\neg\Lovespp\Romeopp{\existsvari y 0}}{} and 
\bigmaths{\Lovespp\Romeopp\Julietpp}. \ 
In practice of \maslong s, 
however, it is often necessary to expand connectionless parts 
to support the speculation of lemmas, which then provide a 
``connection'' that is not syntactically obvious, but closes the
branch nevertheless. This is especially the case for inductive theorem
proving for theoretical \cite{induction-no-cut} and practical 
\nlbcite{mandat,samoacalculemus,samoa-lemmas} reasons.

\yestop\noindent{\bf 
Notational Redundancy} means in a sequent-calculus proof that
the offspring sequents repeat the formulas of 
their ancestor sequents again and again. 
This is partly overcome in the corresponding tableau calculi. 
But even tableau proofs repeat the 
subformulas of their\emph{principal formulas} as\emph{side formulas}
\nlbcite{gentzen} again and again\@.
\emph{Structure sharing} can overcome this redundancy
and does not differ much 
for sequent, tableau, or matrix calculi 
because information on branch, \math\gamma-multiplicity, and fairness 
has to be stored anyway.
As \nolinebreak \maslong s are still
far from delivering what they once promised to achieve,
this optimization is, however, not of top priority,
especially because structure sharing is not trivial, but
likely to block other improvements:
Note that\emph{\math\gamma-step multiplicity} requires variable renaming
and that different rewrite steps may be applied
to the multiple occurrences of subformulas.\footnote
{Indeed, in 
 \cite{isabellesevenhundred} we read:
 \begin{quote}
 ``\ml's execution profiler reported that the sharing mechanism, 
 meant to boost efficiency, was consuming most of the run time.
 The replacement of structure sharing by copying made \ISABELLE\	
 simpler and faster. Complex algorithms are often the problem, 
 not the solution.''\end{quote}}

\yestop\noindent{\bf\NonPermutability} is the subject of this \daspaper. 
Very roughly speaking, it means that the\emph{order} of inference\emph{steps}
(\ie\ applications of reductive inference rules) 
may be crucial for a proof to succeed.
Roughly speaking, permutability of two steps \nlbmath{S_1} and 
\nlbmath{S_0} simply means the following:
{\em In a closed proof tree where 
 \nlbmath{S_0} precedes \nlbmath{S_1}
 and where \nlbmath{S_1} was already applicable before \nlbmath{S_0},
 we can do the step \nlbmath{S_1} before \nlbmath{S_0}
 and find a closed proof tree nevertheless.}
When several formulas in a sequent classify as principal formulas
of \math\alpha-, \math\beta-, \math\gamma-, or \math\delta-steps, 
the search space is typically non-confluent. 
Therefore, 
a bad order of application of these inference steps
may require the search procedure
to backtrack or 
to construct a proof on a higher level of \math\gamma-multiplicity 
than necessary or 
than a mathematician would expect.
Notice that the latter gives a human user 
hardly any chance to cooperate in proof construction:
Who would tell the system to apply a lemma twice when he knows that one 
application suffices?

When we do a \math\gamma-step first and a \math{\delta}-step 
second, a proof may fail on the given level of \math\gamma-multiplicity, 
whereas it succeeds
when we apply the \math{\delta}-step first and the \math\gamma-step second. \ 
For sequent calculi without free variables (\cf\ \eg\ \cite{gentzen}) 
this is exemplified
in \cite[Chapter\,1, \litsectref{4.3.2}]{wallen}. \ 
The reason for this \nonpermutability\ is simply 
that, for the first alternative, 
due to the eigenvariable condition, the \math\gamma-step
cannot instantiate its side formula with the parameter 
introduced by the \math{\delta}-step. \ 

This \nonpermutability\ is not overcome with the introduction of \fev s,
resulting in the so-called 
``\freevariable'' calculi \nlbcite{fitting,wirthcardinal}: \ 
The reason now is that, for the first alternative, 
the\emph\vc\ blocks the \fev\ \nlbmath{\existsvari y{}} 
introduced by the \math\gamma-step %to its side formula
against the instantiation of any term containing 
the\emph\wfuv\ \nlbmath{\wforallvari x{}}
introduced by the \deltaminus-step.
In \nolinebreak\skolemizing\ inference systems, however, 
we would have to say that 
\math{\existsvari y{}} becomes an argument of the \skolem\ term
\app{\wforallvari x{}}{\ldots\existsvari y{}\ldots} introduced by the
\deltaminus-step, which causes unification of 
\existsvari y{} and
\app{\wforallvari x{}}{\ldots\existsvari y{}\ldots}
to fail by the occur check.

This \nonpermutability\ is overcome in \cite[Chapter\,2]{wallen}
with a matrix calculus which generates \vc s equivalent
to\emph{Outer \skolemization}. \ 
\mbox{As a \deltaplus-step \cite{deltaplus}}
extends the \vc\ only equivalently to\emph{Inner \skolemization}
(which is an improvement over Outer \skolemization,
 \ie\ less blockings, or less occurrences in \skolem-terms
 \nlbcite{strongskolem}), 
this \nonpermutability\ is \afortiori\ overcome by the replacement
of the \deltaminus-steps with \deltaplus-steps. \ 

\yestop\noindent{\bf Optimization Problems}
where a badly chosen order of inference steps
does not cause a failure of the proof
(at \nolinebreak the \nolinebreak current level of \math\gamma-multiplicity)
but only an increase in proof size,
are not subsumed under the notion of \nonpermutability.
A typical optimization problem is the following:
The size of a proof crucially depends
on the \math\beta-steps being applied not too early and in the right order.
This is obvious
from a working mathematician's point of view: 
Do not start a case analysis before it is needed and make the nested case 
assumptions in an order that unifies identical argumentations!

\yestop\noindent
Thus, assuming an any-time behavior of a semi-decision procedure
for closedness running in parallel
(\tightemph{simultaneous rigid \math E-unification} is not co-semi-decidable
\nlbcite{simultaneousrigideunification}),
the\nolinebreak\emph{folklore heuristics} is somewhat as follows:
\par\yestop\noindent{\bf Step\,1: }
Apply all \math\alpha- and \math\delta-steps, guaranteeing termination by
deleting their principal
formulas from the child sequents %of the side formulas 
(either directly syntactically in sequent calculi, 
 or indirectly by some bookkeeping for search control in tableau calculi).
\par\yestop\noindent{\bf Step\,2: }
If a \math\gamma-rule is applicable to a principal formula that has not
reached the current threshold for \math\gamma-multiplicity in some branch, 
do such a \math\gamma-step, 
namely the one with the most promising connections, and then go to Step\,1.
\par\yestop\noindent{\bf Step\,3: }
If a \math\beta-rule is applicable, then apply the most promising one, 
deleting  its principal
formula from the sequents of the side formulas, and then go to Step\,1.
Otherwise, if a \math\gamma-rule is applicable,
then increase the threshold for \math\gamma-multiplicity, 
and then go to Step\,2.\vfill\pagebreak

%%%%%%%%%%%%%%%%%%%%%%%%%%%%%%%%%%%%%%%%%%%%%%%%%%%%%%%%%%%%%%%%%%%%%%%%%%%%%%%%
\section{Background Required for the Example Proof}\label{section background}
\yestop\noindent
Before we go on with this abstract expert-style discussion in 
\sectref{section discussion}, 
we do the proof of {\nolinebreak$($$\lim$$+$$)$} in \sectref{section proof}. \ 
To this end, we now present a sub-calculus of the 
calculus of \cite{wirthcardinal}, whose development was
driven by the integration of \fermat's\emph\descenteinfinie\
into state-of-the-art deduction, 
with human-orientedness as the second design goal.
The calculus uses \vc s instead of \skolemization.
\Vc s are isomorphic to \skolem izaton
in the relevant aspects of this \daspaper,
but admit the usage of simple variables instead of huge \skolem\ terms.
This improves the readability of our formal proof significantly.
We \nolinebreak assume the following sets of\emph{variables} to be disjoint:
\par\yestop\noindent\mbox{}\hfill\math{\begin{tabular}{@{}l l@{}}
  \Vsome
 &\tightemph{\fev s},
  \ie\ 
  the free variables of \cite{fitting}
\\\Vall
 &\tightemph{\fuv s},
  \ie\ 
  nullary parameters, instead of \skolem\ functions
\\\Vbound
 &\tightemph{bound variables},
  \ie\ variables to be bound, \cf\ below
\end{tabular}}\par\yestop\noindent
We \nolinebreak 
use `\math\uplus' for the union of disjoint classes.
We partition the \fuv s into\emph{\wfuv s} 
and\emph{\sfuv s}: \bigmaths{\Vall=\Vwall\uplus\Vsall}.
We define the\emph{free variables} by \bigmaths{\Vfree:=\Vsome\uplus\Vall}{}
and the\emph{variables} by \bigmaths{\V:=\Vbound\uplus\Vfree}.
Finally, the\emph{rigid} variables by \bigmaths{
\Vsomesall:=\Vsome\uplus\Vsall}.
We use \bigmath{\VARsingleindex k(\Gamma)}
to denote the set of variables from
\nolinebreak\Vsingleindex k occurring in \nlbmath\Gamma. \ 
We do not permit binding of variables that already occur 
 bound in a term or formula;
 that \nolinebreak is: \bigmath{\forall x\stopq A} is only a formula 
 if no binder on \math x already occurs in \nlbmath A.
 % \mbox{``\math{\forall x\stopq}''}, 
 % \nolinebreak\mbox{``\math{\exists x\stopq}''}, 
 % \nolinebreak\mbox{``\math{\lambda x\stopq}''}, 
 % \nolinebreak\mbox{``\math{\varepsilon x\stopq}''}.
 The simple effect is that our formulas are easier to read
 and
 our \math\gamma- \nolinebreak 
 and \math\delta-rules can replace\emph{all} occurrences of \nlbmath x.
 Moreover, we assume that all binders have 
 minimal scope.

Let \math\sigma\ be a\emph{substitution}.
We say that \math\sigma\ \nolinebreak is a\emph{substitution on \math X} 
\udiff\ \ \mbox{\math{\DOM\sigma\subseteq X}.} \ \ 
We \nolinebreak denote with `\math{\Gamma\sigma}' 
the result of replacing 
each occurrence of a variable \nlbmath{x\in\DOM\sigma} in \nlbmath{\Gamma} 
with \app\sigma x. \ 
Unless otherwise stated,
we tacitly assume that all occurrences of variables from 
\nlbmath\Vbound\ 
in a term or formula or in the range of a substitution
are\emph{bound occurrences} 
(\ie\ \nolinebreak that a variable \math{x\in\Vbound}
occurs only in the scope of a binder on \nlbmath x)
and that each substitution \math\sigma\
satisfies \bigmaths{\DOM\sigma\subseteq\Vfree},
so that no bound occurrences of variables can be replaced 
and no additional variable occurrences can become bound (\ie\ \nolinebreak 
captured) when applying \nlbmath\sigma.

\yestop\begin{definition}[\VC, \math\sigma-Update, \RSub]\label
{definition variable condition}\label{definition update}\label
{definition ex r sub}\label{definition quasi-existential}\mbox{}\\
A\emph\vc\ is a subset of \bigmath{\Vfree\times\Vfree.}
\par\noindent Let \math R be a \vc\ and \math\sigma\ be a substitution.
The\emph{\math\sigma-update of \math R} is 
\par\noindent\LINEmaths{R\quad\cup\quad\setwith
    {\pair{\freevari z{}}{\freevari x{}}}
    {\freevari x{}\tightin\DOM\sigma
     \und
     \freevari z{}\tightin\VARfree{\app\sigma{\freevari x{}}}
    }}.
\par\noindent\math\sigma\ is an\emph{\Rsub}
\udiff\  
\math\sigma\ is a substitution 
and the \math\sigma-update \nlbmath{R'} of \nlbmath R is\emph{wellfounded},
\ie\ for any nonempty set \nlbmath B, there is a \math{b\in B} such that
there is no \maths{a\in B}{} with \bigmaths{a\nottight{R'}b}.\end{definition}

\begin{sloppypar}\yestop\noindent
Note that, regarding syntax, 
\bigmath{\pair{\freevari x{}}{\freevari y{}}\tightin R} 
is intended to mean that an \math R-substitution \nlbmath\sigma\ 
must not replace \nlbmath{\freevari x{}} with a term in which 
\math{\freevari y{}} could ever occur. 
This is guaranteed when
the \math\sigma-updates \nlbmath{R'} of \nlbmath R are always required to
be wellfounded.
Indeed, for \math{\freevari z{}\in\VARfree{\app\sigma{\freevari x{}}}}, \ 
we get \bigmaths{
\freevari z{}\nottight{R'}\freevari x{}\nottight{R'}\freevari y{}},
blocking \nlbmath{\freevari z{}} against terms containing \freevari y{}. \ 
In practice, a \math\sigma-update of \nlbmath R
can always be chosen to be finite. In this case, it \nolinebreak
is wellfounded \uiff\
\mbox{it is acyclic}.\vfill\pagebreak\end{sloppypar}

\thereductiverules h

\subsection{Inference Rules for Reduction Within a Proof Tree}\label
{section inference rules}In \figuref{figure reductive rules},
the inference rules for
reductive reasoning within a tree 
are presented in sequent style.
Note that in the good old days when trees grew upwards, \gentzen\ 
would have inverted the inference rules 
such that passing the line means consequence. 
In \nolinebreak our case, passing the line means reduction,
and trees \mbox{grow downwards}.

All rules are\emph{sound}
and\emph{solution preserving} for the rigid variables
in the sense of \mbox{\cite[\litsectref{2.4}]{wirthcardinal}}. \ 
Thus, updating a global \vc\ \nlbmath R, 
we can globally apply any \math R-substitution on any subset of \nlbmath\Vsome\ 
without destroying the soundness %and safeness 
of the instantiated proof steps. 

Instead of an eigenvariable condition, the \deltaminus-rules
come with a binary relation on variables to the lower right, which
must be added to the current \vc\ \nlbmath R\@. \ 
The \deltaplus-rules come with an additional relation to the upper right,
which has to be added to the\emph{\math R-\cc} \nlbmath C.
This \cc\ is an optional part of the calculus.
It \nolinebreak 
may store a structure-sharing
representation of an \math\varepsilon-term \cite
{grundlagen,ahrendtgiese,wirthhilbertepsilon}
for a \sfuv,
which may restrict the possible values of this variable.
As they play only a marginal role in the example proof of \sectref
{section proof}, we do not have to discuss \cc s here.
Note, however, that without a \cc,
the \deltaplus-rules would only be sound but not solution preserving,
\cf\ \examref{example eta}.

Indeed, the calculus contains\emph{different kinds of \math\delta-rules in
parallel}. There\-fore---to be sound---the 
\deltaminus-rules have to refer to the the
\sfuv s introduced by the \deltaplus-rules in their \vc s, and vice versa.

\subsection{Lemma Application Between Proof Trees}\label
{section lemma application}The reason why we spoke of a proof\emph{forest} 
\math{\mathcal F} in \figuref{figure reductive rules} is 
that a proof may be spread over several trees that are connected by generative 
application of the root of one tree in the reductive proof of another tree, 
either as a lemma or as an induction hypothesis. While the application
of lemmas must be wellfounded, induction hypotheses may be applied to the
proof of themselves and mutually. 
In this \daspaper, we only need lemma application. 

Lemma application works as follows.
When a lemma \bigmath{A_1,\ldots,A_m} 
is a subsequent of a leaf sequent \nlbmath\Gamma\ to be proved
(\ie\ if, for all \math{i\in\{1,\ldots,m\}}, 
 the formula \nlbmath{A_i} is listed in \nlbmath\Gamma), 
its application 
closes the branch of this sequent ({\em subsumption}). \ 
Otherwise, the conjugates of the missing formulas \nlbmath{C_i}
are added to the child sequents (premises), one child per missing formula. 
This can be seen as Cuts on \nlbmath{C_i} plus subsumption.
More precisely---modulo associativity, commutativity, and idempotency---a 
sequent 
\bigmath{A_1,\ldots,A_m,B_1,\ldots,B_n}
can be reduced by application of the lemma
\bigmath{A_1,\ldots,A_m,C_1,\ldots,C_p} to the sequents 
\par\yestop\noindent\LINEnomath{
\bigmaths{\overline{C_1},A_1,\ldots,A_m,B_1,\ldots,B_n}{}
\hfill\math\cdots\hfill\bigmaths{\overline{C_p},A_1,\ldots,A_m,B_1,\ldots,B_n}.
}\par\yestop\noindent
In addition, any time we apply a lemma, we can replace its \wfuv s 
locally and arbitrarily,
except those \wfuv s that depend on rigid variables which (in rare cases)
may already occur in the input lemma. 
More precisely, the set of \wfuv s of a lemma
\nlbmath\Phi\ we may instantiate is exactly
\par\yestop\noindent\LINEmaths{
  \displaysetwith
    {\wforallvari y{}\tightin\VARwall\Phi}
    {\VARsomesall\Phi\times\{\wforallvari y{}\}\nottight{\nottight{\subseteq}}R
    }}.\par\yestop\noindent
Typically \VARsomesall\Phi\nolinebreak\ is empty and no restrictions apply.
Note that we also may extend this set of \wfuv s
by extending the \vc\ \nlbmath R.
This instantiation of outermost \deltaminus-variables mirrors mathematical
practice, saves repetition of initial \math\delta-steps, and is essential 
for induction, 
where the weights depend on these \wfuv s to guarantee wellfoundedness.
There will be a sufficient number of self-explanatory 
examples of application of\emph{open lemmas} (\ie\ yet unproved lemmas) 
in \sectref{section proof}.

\vfill\pagebreak\figureglobaldefinitions h\pagebreak

\section{The \protect\math{(\protect\lim\protect\tight+)} Proof: 
 Limit Theorem on Sums in \RE}\label
{section proof}\subsection{Explanation and Initialization}\label
{section explanation and initialization}
Compared to the proof of {\nolinebreak$($$\lim$$+$$)$} as presented in 
the lecture courses,
the version we present here admits 
a more rigorous argumentation for \nonpermutability\ 
of \math\beta\ and \deltaplus\ in the following sections.\footnote
{I did not succeed in finding a really satisfying definition of non-local 
 permutability
 that fits the non-local situation of the failure of the 
 \math{(\protect\lim\protect\tight+)} proof as presented in the lecture 
 \mbox{courses \cite{maslecture,wirthlecture}}. \
 The problem was to
 permute the critical \math\beta-step from below 
 the critical \mbox{\deltaplus-steps}
 to a place far up above the \deltaplus-steps. \ 
 And on this partial path 
 from \math\beta\ down to \deltaplus\
 there were other inference 
 steps which may or may not contribute to the non-permutability. \ 
 Thus, instead of globalizing the notion of permutability I
 localized the example proof; \ 
 although the original version had pedagogical advantages.
 \par
 Furthermore, note that 
 it may be possible to demonstrate the permutability problems of
 the \math\beta-rule with slightly smaller artificial examples.
 But we prefer a practical example to demonstrate the practical difficulties
 and discuss some less formal soft aspects which may be more important than
 the hard \nonpermutability\ results of this \daspaper.
 Moreover, because of its many interesting aspects, 
 this proof will be useful as a standard example for further 
 reference. 
 If you are not in love with formal proofs, I do apologize for the 
 inconvenience of my decision and ask you to send me an 
 \EMAIL\ of complaint if you will not have learned something that is worth
 your efforts in the end. 
 If I receive at least three \EMAIL s seriously
 stating that these efforts were in vain but the \nonpermutability\
 deserves proper publication,
 I will try to produce a version of this \daspaper\ with a somewhat smaller
 artificial example.}

\noindent
By standard mathematical abuse of notation, 
we want to prove the theorem\footroom
\par\noindent\math{(\lim\tight+)}\LINEmath
{\displaystyle\lim_{x\rightarrow\wforallvari x 0}
    \inparentheses{\app{\wforallvari f{}}x+\app{\wforallvari g{}}x}
        =\displaystyle\lim_{x\rightarrow\wforallvari x 0}\app{\wforallvari f{}}x
         +
         \displaystyle\lim_{x\rightarrow\wforallvari x 0}\app{\wforallvari g{}}x
}

\noindent
Before we start the formal proof, we expand {\nolinebreak$($$\lim$$+$$)$} 
into a better notation:
\par\yestop\noindent(1): \hfill\math{
    \inparenthesesoplist{
        \displaystyle\lim_{x\rightarrow\wforallvari x 0}\app{\wforallvari f{}}x
        =\wforallvari y f
      \oplistund
        \displaystyle\lim_{x\rightarrow\wforallvari x 0}\app{\wforallvari g{}}x
        =\wforallvari y g
    }
  \implies
    \displaystyle\lim_{x\rightarrow\wforallvari x 0}
    \inparentheses{\app{\wforallvari f{}}x\tight+\app{\wforallvari g{}}x}
        =\wforallvari y f\tight+\wforallvari y g
}\par\yestop\noindent
Warning: The ``\math='' here is still no real equality symbol! 
What is it, then?
Something like
\ \mbox{\maths
{\displaystyle\lim_{x\rightarrow 0}\inparentheses{x^2\sin{1\over x}}=0},} 
\ formally say
\bigmaths{\displaystyle\lim_{x\rightarrow z}t_x=t'}{}
(\tightemph{definiendum}), \  
is defined by the formula (\tightemph{definiens})
\\\LINEmath{
    \forall\varepsilon\tight>0\stopq
    \exists\delta\tight>0\stopq
    \forall x\boldunequal z\stopq
    \inparentheses{
        |t_x\tight-t'|<\varepsilon 
      \nottight\antiimplies
        |x\tight-z|<\delta
    }}\par\noindent
Note that 
\bigmath{\forall\varepsilon\tight>0\stopq A} and 
\bigmath{\exists\delta\tight>0\stopq B} and
\bigmath{\forall x\boldunequal z\stopq C}
(\tightemph{definienda})
abbreviate \ \mbox{\math
{\forall\varepsilon\stopq \inpit{0\tight<\varepsilon\implies A}}} \ 
and 
\bigmath{\exists\delta\stopq \inpit{0\tight<\delta\und B}}
and 
\bigmaths{\forall x\stopq \inpit{x\boldunequal z\implies C}}{}
(\tightemph{definientia}),
respectively.
Thus, if---\inthesequel---we speak of an\emph
{expansion of \ \mbox{``\math{\,\forall\varepsilon\tight>0\stopq\ldots}''}} \ 
(from\emph{definiendum} to\emph{definiens}) \ or simply of\emph
{an expansion of \nlbmath{\,\forall}}, \ we mean the replacement
of \bigmaths{\forall\varepsilon\tight>0\stopq A}{} with
\ \mbox{\math{\forall\varepsilon\stopq \inpit{0\tight<\varepsilon\implies A}}} \
for some formula \nlbmath A \ in a reductive proof step. \ 
Analogous proof steps are meant by\emph{expansion of \math{\,\exists}}
\ and\emph{expansion of \nlbmath{\,\lim}}, \ respectively. \ 
We will often reorder the formulas in the sequents without mentioning it.

\noindent
We initialize our global \vc\ \nlbmath R \ by \ \nlbmath{R:=\emptyset}, \  
and our global   \math R-\cc\ \nlbmath C \ by \ \nlbmath{C:=\emptyset}.%

\subsection{Expanding the Proof Tree with Root \protect\math{(1)}}\label
{section expanding the proof}
By two \math\alpha-steps and expansion of \math\lim\ from\emph{definiendum} 
to\emph{definiens}, we reduce \nlbmath{(1)} to its single child \nlbmath{(1.1)},
writing \math{(1^2)} for \math{(1.1)}:
\par\noindent\math{(1^2)}: \hfill\math{
    \forall\varepsilon\tight>0\stopq 
    \exists\delta\tight>0\stopq 
    \forall x\boldunequal\wforallvari x 0\stopq 
    \inparenthesesoplist{
        |\inpit{\app{\wforallvari f{}}{x}\tight+\app{\wforallvari g{}}{x}}
         -\inpit{\wforallvari y f\tight+\wforallvari y g}|<\varepsilon
      \oplistantiimplies
        |x\tight-\wforallvari x 0|<\delta
    }
  \comma}
\\\mbox{}\hfill\math{
     \displaystyle\lim_{x\rightarrow\wforallvari x 0}\app{\wforallvari f{}}x
     \neq\wforallvari y f
  \comma
     \displaystyle\lim_{x\rightarrow\wforallvari x 0}\app{\wforallvari g{}}x
     \neq\wforallvari y g
}

\noindent
By expansion of 
\ ``\math{\forall\varepsilon\tight>0\stopq\ldots}'' \ from\emph{definiendum} 
to\emph{definiens},
then a 
\mbox{\deltaminus- and an} \math\alpha-step, and finally expansion of 
\nlbmath\exists\ and some reordering of the listed formulas we reduce this to:
\par\noindent\math{\mathbf{(1^3)}}: \ 
\math{
    \exists\delta\stopq\inparentheses{\mediumheadroom
    0\tight<\delta
    \und
    \forall x\boldunequal\wforallvari x 0\stopq 
    \inparenthesesoplist{
        |\inpit{\app{\wforallvari f{}}{x}\tight+\app{\wforallvari g{}}{x}}
         -\inpit{\wforallvari y f\tight+\wforallvari y g}|
         <\wforallvari\varepsilon{}
      \oplistantiimplies
        |x\tight-\wforallvari x 0|<\delta}}
  \comma
}\\\mbox{}\hfill\math{
    0\tight\nless\wforallvari\varepsilon{}\tight
    \comma
     \displaystyle\lim_{x\rightarrow\wforallvari x 0}\app{\wforallvari f{}}x
     \neq\wforallvari y f
  \comma
     \displaystyle\lim_{x\rightarrow\wforallvari x 0}\app{\wforallvari g{}}x
     \neq\wforallvari y g
}

\begin{slide}
A \math{\gamma}-step yields:
\par\yestop\noindent\math{\mathbf{(1^4)}}: 
\hfill\math{\criticalbetaformula\comma(1^3)
}\par\yestop\noindent
Note that the \math{(1^3)} at the end of the sequent \nlbmath{(1^4)}
means that the whole parent sequent is part of the child sequent.
\end{slide}

\begin{slide}
Expanding \math\lim\ and \math\forall, plus a \math\gamma-step,
each twice, we get 
(\cfnlb\ \figuref{figure global abbreviations} for \nlbmath\Xi):
\par\halftop\noindent\math{\mathbf{(1^5)}}:
\hfill\math{\begin{array}[t]{@{}l@{}}
    \neg
    \inparentheses{
    0\tight<\existsvari\varepsilon f
    \implies
    \exists\delta_f\tight>0\stopq 
    \forall x_f\boldunequal\wforallvari x 0\stopq 
    \inparenthesesoplist{
         |\app{\wforallvari f{}}{x_f}\tight-\wforallvari y f|
        <\existsvari\varepsilon f
      \oplistantiimplies
        |x_f\tight-\wforallvari x 0|<\delta_f
    }}
  \comma
\\\neg
    \inparentheses{
    0\tight<\existsvari\varepsilon g
    \implies
    \exists\delta_g\tight>0\stopq 
    \forall x_g\boldunequal\wforallvari x 0\stopq 
    \inparenthesesoplist{
         |\app{\wforallvari g{}}{x_g}\tight-\wforallvari y g|
        <\existsvari\varepsilon g
      \oplistantiimplies
        |x_g\tight-\wforallvari x 0|<\delta_g
    }}
  \comma
  \Xi
\\\end{array}}\end{slide}
\begin{slide}A \math\beta-step and an expansion of \math\exists, 
each twice, yield:
\par\yestop\noindent\math{\mathbf{(1^5.1)}}:
\hfill\math{\begin{array}[t]{@{}l@{}}
  0\tight<\existsvari\varepsilon f\comma
    \neg
    \inparentheses{
    0\tight<\existsvari\varepsilon g
    \implies
    \exists\delta_g\tight>0\stopq 
    \forall x_g\boldunequal\wforallvari x 0\stopq 
    \inparenthesesoplist{
         |\app{\wforallvari g{}}{x_g}\tight-\wforallvari y g|
        <\existsvari\varepsilon g
      \oplistantiimplies
        |x_g\tight-\wforallvari x 0|<\delta_g
    }}
  \comma
  \Xi
\\\end{array}}\par\noindent\math{\mathbf{(1^5.2)}}:
\hfill\math{\begin{array}[t]{@{}l@{}}
  0\tight<\existsvari\varepsilon g\comma
    \neg\exists\delta_f\tight>0\stopq 
    \forall x_f\boldunequal\wforallvari x 0\stopq 
    \inparenthesesoplist{
         |\app{\wforallvari f{}}{x_f}\tight-\wforallvari y f|
        <\existsvari\varepsilon f
      \oplistantiimplies
        |x_f\tight-\wforallvari x 0|<\delta_f
    }
    \comma
    \Xi
\\\end{array}}\par\noindent\math{\mathbf{(1^5.3)}}:
\mbox{}\hfill\math{\begin{array}[t]{@{}l@{}}
  \oplistprincipalformuladeltaplusnegexistsdelta f
  \comma
\\\oplistprincipalformuladeltaplusnegexistsdelta g\comma
  \Xi
\\\end{array}}\end{slide}

\begin{slide}%
A \deltaplus-step applied to the first formula at \math{(1^5.3)} yields:
\par\noindent\math{\mathbf{(1^5.3.1)}}:
\mbox{}\hfill\math{
    0\tight<\existsvari\delta{}    
    \und
    \forall x\boldunequal\wforallvari x 0\stopq 
    \inparenthesesoplist{
        |\inpit{\app{\wforallvari f{}}{x}\tight+\app{\wforallvari g{}}{x}}
         \tight-\inpit{\wforallvari y f\tight+\wforallvari y g}|
        <\wforallvari\varepsilon{}
      \oplistantiimplies
        |x\tight-\wforallvari x 0|<\existsvari\delta{}
    }\comma\Theta}\par\noindent
where \nlbmath R is extended with
\ \ \bigmaths{\vcdeltaplusnegexistsdelta f}, \ \ and the \cc\ \nlbmath C with:
\\\noindent\mbox{}\hfill
{\displayset{\oplistccdeltaplusnegexistsdelta f}}
\end{slide}

\yestop\subsection{A Bad Turn}\label{section bad turn}
Now we do an early \math\beta-step against the folklore 
heuristics presented in \sectref{section introduction}.
This will make the whole following subproof fail!
A \nolinebreak reader who is interested only in a successful example proof 
may continue reading with \sectref{section backtracking}.
\par\yestop\noindent\math{\mathbf{(1^5.3.1.1)}}:
\hfill\math{
    0\tight<\existsvari\delta{}\comma\Theta
}\par\noindent\math{\mathbf{(1^5.3.1.2)}}:
\hfill\math{
    \forall x\boldunequal\wforallvari x 0\stopq 
    \inparenthesesoplist{
        |\inpit{\app{\wforallvari f{}}{x}\tight+\app{\wforallvari g{}}{x}}
         -\inpit{\wforallvari y f\tight+\wforallvari y g}|<
         \wforallvari\varepsilon{}
      \oplistantiimplies
        |x\tight-\wforallvari x 0|<\existsvari\delta{}
    }
    \comma\Theta}
\vfill\pagebreak

\yestop\noindent A \deltaplus-step,
two \math\alpha-steps, and expansion of \math\forall, 
applied to \nlbmath{(1^5.3.1.2)}, yield:
\par\noindent\math{\mathbf{(1^5.3.1.2.1)}}:
\mbox{}\hfill\math{
    \forall x\stopq 
    \inparentheses{
    x\boldunequal\wforallvari x 0
    \implies
    \inparenthesesoplist{
        |\inpit{\app{\wforallvari f{}}{x}\tight+\app{\wforallvari g{}}{x}}
         -\inpit{\wforallvari y f\tight+\wforallvari y g}|
        <\wforallvari\varepsilon{}
      \oplistantiimplies
        |x\tight-\wforallvari x 0|<\existsvari\delta{}
    }}
  \comma\Omega}\par\noindent
where \nlbmath R is extended with
\ \ \bigmaths{\vcdeltaplusnegexistsdelta g}, \ \ and \nlbmath C with:
\\\mbox{}\hfill
{\displayset{\oplistccdeltaplusnegexistsdelta g}}

\begin{slide}A \deltaplus-step and
two \math\alpha-steps yield
(\cf\ \figuref{figure global abbreviations} for \nlbmath t):
\par\noindent\math{{(1^5.3.1.2.1^2)}}:
\hfill\math{
\lastlineoffifthcase
}\par\noindent
where \nlbmath R is extended with
\bigmath{\{
\wforallvari x 0,
\wforallvari f{},
\wforallvari g{},
\wforallvari y f,
\wforallvari y g,
\wforallvari\varepsilon{},
\existsvari\delta{}
    \}\times\{\sforallvari x{}\}}\\
and our \math R-\cc\ \nlbmath C with 
\par\noindent\mbox{}\hfill 
\math{\displayset{\sforallvari x{}\mapsto\neg\inparentheses{
    {\sforallvari x{}}\boldunequal\wforallvari x 0
    \implies
    \inparentheses{t<\wforallvari\varepsilon{}
      \nottight{\nottight\antiimplies}
        |{\sforallvari x{}}\tight-\wforallvari x 0|<\existsvari\delta{}
    }}}}
\end{slide}\begin{slide}Expansion of \math\forall\
and a \math{\gamma}-step, each twice, yield:
\par\noindent\math{(1^5.3.1.2.1^3)}:
\hfill\math{\protofifthcase}\end{slide}

\subsection{Partial Success}\label
{section partial success}\figurefailedproof t

2 \math\beta-steps,
each twice, yield:
\par\yestop\noindent\math{(1^5.3.1.2.1^3.1)}: \ 
\math{\existsvari x f\boldunequal\wforallvari x 0\comma
    \sforallvari x{}\boldequal\wforallvari x 0
    \comma\ldots
}
\par\noindent\math{(1^5.3.1.2.1^3.2)}: \ 
\math{\existsvari x g\boldunequal\wforallvari x 0\comma
    \sforallvari x{}\boldequal\wforallvari x 0
    \comma\ldots
}
\par\noindent\math{(1^5.3.1.2.1^3.3)}: \ 
\math{
    |\existsvari x f\tight-\wforallvari x 0|<\sforallvari\delta f
  \comma
     |\sforallvari x{}\tight-\wforallvari x 0|\nless\existsvari\delta{}
  \comma\ldots
}
\par\noindent\math{(1^5.3.1.2.1^3.4)}: \ 
\math{
    |\existsvari x g\tight-\wforallvari x 0|<\sforallvari\delta g
  \comma
     |\sforallvari x{}\tight-\wforallvari x 0|\nless\existsvari\delta{}
  \comma\ldots
}
\par\noindent\math{(1^5.3.1.2.1^3.5)}:
\ \mbox{}\fifthcase

\begin{slide}
And now?
By formula unification and some basic knowledge of the domain, 
we can easily see that global application of the substitution \nlbmath\sigma\
from \sectref{section explanation and initialization}
admits to close the branches of the first four sequents.
According to \defiref{definition update}, this adds \\\linemaths{\{ 
\pair{\sforallvari x{}}{\existsvari x f},
\pair{\sforallvari x{}}{\existsvari x g},
\pair{\sforallvari\delta f}{\existsvari\delta{}},
\pair{\sforallvari\delta g}{\existsvari\delta{}}\}}{} to our \vc\ \nlbmath R,
which, luckily, stays acyclic, \cf\ the 
acyclic graph of \figuref{figure variable-condition one} in
\sectref{section where variable condition after sigma is}. \ 
\math{(1^5.3.1.2.1^3.1)} and \math{(1^5.3.1.2.1^3.2)} become logical axioms.
Applying lemma \nlbmath{(2)} 
of \figuref{figure global abbreviations} instantiated via \math{\{
    \wforallvari y{}\tight\mapsto\sforallvari\delta f
  \comma
    \wforallvari z{}\tight\mapsto\sforallvari\delta g
\}} we reduce \math{(1^5.3.1.2.1^3.3)} to:
\par\yestop\noindent\math{(1^5.3.1.2.1^3.3.1)}:
\ \math{
     \min(\sforallvari\delta f,\sforallvari\delta g)\nleq
     \sforallvari\delta f
  \comma
     |\sforallvari x{}\tight-\wforallvari x 0|<\sforallvari\delta f
  \comma
%}\\\mbox{}\hfill\math{
     |\sforallvari x{}\tight-\wforallvari x 0|\nless
     \min(\sforallvari\delta f,\sforallvari\delta g)
  \comma
  \ldots
}\par\yestop\noindent
which is subsumed by the transitivity lemma \nlbmath{(3)}
of \figuref{figure global abbreviations}.
\\\math{(1^5.3.1.2.1^3.4) }
can be closed analogously to \nlbmath{(1^5.3.1.2.1^3.3)}.
\end{slide}

\subsection{Total Failure}\label{section total failure}
Abstractly, our proof tree looks as in \figuref{figure failure}.
By the application of \nlbmath\sigma, \ \math{(1^5.3.1.1)} has become
\\\LINEmaths{
  0\tight<\min(\sforallvari\delta f,\sforallvari\delta g)\comma\Theta
}{}\par\halftop\noindent
If \nolinebreak the \nolinebreak first formula---which is the only
new one as compared to its parent sequent---is irrelevant for the proof 
of \nlbmath{(1^5.3.1.1)} (in the sense that it is not contributing
as a principal formula, \cf\ \cite{gentzen, mandat, samoa-lemmas}),
then we had better prove \math{(1^5.3.1)} instead, because this saves us the
proof of the whole \math{\beta_2}-subtree of \nlbmath{(1^5.3.1)}.
But look:
\sforallvari\delta g \nolinebreak is  \nolinebreak not introduced 
before \math{(1^5.3.1.2.1)}, 
which in \math{(1^5.3.1.2.1^2)} results in the context
\bigmaths{0\tight\nless\sforallvari\delta f
    \comma
      0\tight\nless\sforallvari\delta g}{}
(as listed in \nlbmath\Omega\ of \figuref{figure global abbreviations})
with which we could prove 
\bigmaths{0\tight<\min(\sforallvari\delta f,\sforallvari\delta g)}{}
by lemma \nlbmath{(4)} of \figuref{figure global abbreviations}. \ 
Thus, the \math\beta-step applied to \math{(1^5.3.1)} does not have any benefit
unless it is done\emph{below} \nlbmath{(1^5.3.1.2.1)}.
\pagebreak

\yestop
\noindent
Now, we have three possibilities in principle:\begin{enumerate}
\yestop\item\label{item backtrack}
We can backtrack to \math{(1^5.3.1)}, deleting all its sub-trees.
\yestop\item We could try to 
use the \cc\ of \sforallvari\delta g
to find out that it is positive. \ 
\ \app C{\sforallvari\delta g} \nolinebreak is
\par\halftop\noindent\LINEmaths{
\kernelofsideformuladeltaplusnegexistsdelta g}.\par\halftop\noindent
But this guarantees \bigmath{0\tight<\sforallvari\delta g} only 
if also the second part of the conjunction can be shown to be satisfiable,
for which we again lack the context.
\yestop\item 
We can prove \nlbmath{(1^5.3.1.1)} by proving its subsequent \nlbmath\Theta. \
As \nlbmath\Theta\ is already a subsequent of \nlbmath{(1^5.3.1)},
this means that we could prove already \nlbmath{(1^5.3.1)} this way. \ 
Thus, the whole subproof below \math{(1^5.3.1.2)} could be pruned. \ 
Moreover, 
as we would have to expand
the principal \math\gamma-formula of \nlbmath{(1^3)} a second time,
resulting in a higher maximum of 
\mbox{\math\gamma-multiplicity} than necessary,
the following lemma holds.\yestop\end{enumerate}

\yestop\begin{lemma}\label{lemma no completion} \ 
Using the reductive rules of \figuref{figure reductive rules}
with a \mbox{\math\gamma-multiplicity} threshold of\/ \nlbmath 1, \
the \nolinebreak 
current proof tree (with the partial instantiation \nlbmath\sigma) 
cannot be expanded and instantiated 
to a closed proof tree at \math{(1^5.1)}, \math{(1^5.2)}, and 
\nlbmath{(1^5.3.1.1)} in parallel.\end{lemma}

\yestop\noindent
For a proof of \lemmref{lemma no completion} 
\cfnlb\ \sectref{section proof of lemma}. \ 
Note that the validity of \lemmref{lemma no completion} depends
on the \deltaminus- and \deltaplus-rules being the only \math\delta-rules
available. With \deltaplusplus-rules the situation would be different,
\cfnlb\ \sectref{section deltaplusplus}. \ 
Moreover, as our proof trees are customary AND-trees 
(and no AND/OR-trees that admit alternative proof attempts 
 as in \nlbcite{pds,quodlibet-cade}), \lemmref{lemma no completion}
means that the whole proof attempt is failed for a 
\mbox{\math\gamma-multiplicity} of \nlbmath 1.

\yestop\yestop
\subsection{Backtracking to the Path of Virtue}\label{section backtracking}

\begin{sloppypar}
Item\,\ref{item backtrack} in the above list is the only reasonable alternative.
Therefore, let \nolinebreak us restart from 
\nlbmath{(1^5.3.1)} ---not without storing \nlbmath\sigma\
and its connections before.\end{sloppypar}

\yestop\noindent
Applied to \nlbmath{(1^5.3.1)},
one \deltaplus-step,
two \math\alpha-steps,
two expansions of \nlbmath\forall, and two \math\gamma-steps
yield as in \sectref{section bad turn} and with the same extensions of 
\nlbmath R and \nlbmath C:
\par\noindent\math{(1^5.3.1^2)}:
\hfill\math{\begin{array}[t]{@{}l@{}}
  \neg\inparentheses{
    \oplistexpandedsecondhalfvarilesskernelofsideformuladeltaplusnegexistsdelta
    f\sforallvari}
  \comma
\\\neg\inparentheses{
    \oplistexpandedsecondhalfvarilesskernelofsideformuladeltaplusnegexistsdelta
    g\sforallvari}\comma
\\\criticalbetaformula\comma\Omega
\\\end{array}}

\vfill\pagebreak

\figurevision h

\yestop\yestop\noindent
Now we\emph{have to} expand one of the three 
first \math\beta-formulas of \math{(1^5.3.1^2)}.
Note that the third one is the one whose expansion made our proof fail before.
We have learned that the path of virtue is narrow! 
What about taking the first \math\beta-formula? 
This would result in the subtree depicted in \figuref{figure vision} above!
Its first \math\beta-step can
represent progress only if the first \mbox{(\math{\beta_1}-) child}
is easier to prove than the root itself. 
But the only reasonable connection of its single new formula 
\bigmaths{\framebox{\framebox{\math{
\existsvari x f\boldunequal\wforallvari x 0}}}}{}
is to the third
formula \bigmath{\framebox{\framebox{\math{
  \sforallvari x{}\boldequal\wforallvari x 0}}}} of the rightmost
leaf; via \nlbmath\sigma. \ 
Thus, we would have to copy the proof starting below the second 
\mbox{(\math{\beta_2}-) child} of the root to its first 
\mbox{(\math{\beta_1}-) child}.
But, if we do so, this proof will fail again, due to the following reason:
To close the copied subproof we need the connection between 
the fourth formula 
\bigmaths{{\framebox{\math{
|\sforallvari x{}\tight-\wforallvari x 0|\nless\existsvari\delta{}}}}}{}
of the rightmost leaf and the positive subformula 
\bigmaths{\framebox{\math{
|\existsvari x f\tight-\wforallvari x 0|<\sforallvari\delta f}}}{}
of the formula \nlbmath A;
via \nlbmath\sigma, \nlbmath{(2)}, and \nlbmath{(3)} as at the end of
\sectref{section partial success}. \ 
But this connection is only available at the original position and not 
at the position the subproof is copied to,
because the positive subformula 
is part of the \math{\beta_2}-side formula \nlbmath A
of the \math\beta-step at the root.
All in all, this shows that expanding the first \mbox{\math\beta-formula} of  
\nlbmath{(1^5.3.1^2)} leads to a failure of the proof on the current 
threshold for \math\gamma-multiplicity again.
By \nolinebreak symmetry, the same holds for the
second. 
Thus, we take the third. 
Notice that the \math\beta-step we\emph{have to}
do now is the one whose too early application made
us backtrack before.
\vfill\pagebreak

\yestop\noindent
A \math\beta-step to the third \math\beta-formula of 
\nlbmath{(1^5.3.1^2)}, and
expansion of \math\forall\ yield:
\par\halftop\noindent\math{(1^5.3.1^2.1)}:
\ \math{0\tight<\existsvari\delta{}\comma
   0\tight\nless\sforallvari\delta f
 \comma
   0\tight\nless\sforallvari\delta g\comma\ldots}
\par\noindent\math{(1^5.3.1^2.2)}:
\hfill\math{\begin{array}[t]{@{}l@{}}
  \neg\inparentheses{
\oplistexpandedsecondhalfvarilesskernelofsideformuladeltaplusnegexistsdelta
f\sforallvari}\comma
\\\neg\inparentheses{
\oplistexpandedsecondhalfvarilesskernelofsideformuladeltaplusnegexistsdelta
g\sforallvari}\comma
\\\forall x\stopq 
    \inparentheses{
    x\boldunequal\wforallvari x 0
    \implies
    \inparenthesesoplist{
        |\inpit{\app{\wforallvari f{}}{x}\tight+\app{\wforallvari g{}}{x}}
         \tight-\inpit{\wforallvari y f\tight+\wforallvari y g}|
        <\wforallvari\varepsilon{}
      \oplistantiimplies
        |x\tight-\wforallvari x 0|<\existsvari\delta{}
    }}
  \comma\Omega
\\\end{array}}

\noindent
As a \deltaminus-step with the first formula of the last line of 
\nlbmath{(1^5.3.1^2.2)}
as principal formula
would block the later 
instantiation of \existsvari x f
and \existsvari x g with the newly introduced \fuv,
for the proof to succeed on the current 
threshold for \nlbmath\gamma-multiplicity,
we have to take a \deltaplus-step instead.
Note that this was not yet a problem for the sequent 
\nlbmath{(1^5.3.1.2.1)} of \sectref{section bad turn}, 
in which \existsvari x f and \existsvari x g did not occur yet.
Besides the \deltaplus-step extending 
\math R and \math C as in \sectref{section bad turn}, 
we do two \math\alpha-steps.
This results exactly in what was 
seen before at the end of \sectref{section bad turn},
with the exception of a different label:
\par\noindent\math{(1^5.3.1^2.2.1)}:
\hfill\math{\protofifthcase}

\begin{slide}
Again, two \nlbmath\beta-steps, each twice, yield:
\par\yestop\noindent\math{(1^5.3.1^2.2.1.1)}: 
\ \math{\existsvari x f\boldunequal\wforallvari x 0\comma
    \sforallvari x{}\boldequal\wforallvari x 0
    \comma\ldots
}\par\noindent\math{(1^5.3.1^2.2.1.2)}: 
\ \math{\existsvari x g\boldunequal\wforallvari x 0\comma
    \sforallvari x{}\boldequal\wforallvari x 0
    \comma\ldots
}\par\noindent\math{(1^5.3.1^2.2.1.3)}:
\ \math{
    |\existsvari x f\tight-\wforallvari x 0|<\sforallvari\delta f
  \comma
     |\sforallvari x{}\tight-\wforallvari x 0|\nless\existsvari\delta{}
  \comma\ldots
}\par\noindent\math{(1^5.3.1^2.2.1.4)}:
\ \math{
    |\existsvari x g\tight-\wforallvari x 0|<\sforallvari\delta g
  \comma
     |\sforallvari x{}\tight-\wforallvari x 0|\nless\existsvari\delta{}
  \comma\ldots
}\par\noindent\math{(1^5.3.1^2.2.1.5)}:
\ \mbox{}\fifthcase
\end{slide}

\begin{slide}
As before in \sectref{section partial success},
application of \nlbmath\sigma\ admits the closure of
of the four branches of  
\math{(1^5.3.1^2.2.1.[1\mbox{--}4])}.
But now, contrary to what made us backtrack before, 
\math{(1^5.3.1^2.1)} becomes
\par\noindent\LINEmaths{0\tight<\min(\sforallvari\delta f,\sforallvari\delta g)
 \comma
   0\tight\nless\sforallvari\delta f
 \comma
   0\tight\nless\sforallvari\delta g\comma\ldots},\par\noindent
which is subsumed by an instance of lemma \nlbmath{(4)}
of \figuref{figure global abbreviations}.
\end{slide}\vfill\pagebreak

\subsection{A Working Mathematician's Immediate Focus}\label
{section immediate focus}
Note that \math{(1^5.3.1^2.2.1.5)} 
would have been the immediate focus of a working mathematician.
He would have sequenced all the lousy \math\beta-steps\emph{after} doing the
crucial steps of the proof which we can do only now.\emph{Notice that the matrix
(indexed formula tree)
versions of our calculus will enable us to support this human behavior in the 
follow-up lectures.}
Let us repeat \math{(1^5.3.1^2.2.1.5)} with some omissions and some reordering:
\par\halftop\noindent\LINEmath{t<\wforallvari\varepsilon{}\comma
        |\app{\wforallvari f{}}{\sforallvari x{}}\tight-\wforallvari y f|
        \nless\existsvari\varepsilon f
  \comma
        |\app{\wforallvari g{}}{\sforallvari x{}}\tight-\wforallvari y g|
        \nless\existsvari\varepsilon g
  \comma
  \ldots}
\par\halftop\noindent
where \bigmath{t<\wforallvari\varepsilon{}} 
actually reads (with some added wave-front annotation to 
be used in \sectref{section clean up})
\par\halftop\noindent\LINEmath{
 |\,\,
  \framebox{\math{\inpit{\underline{\app{\wforallvari f{}}{\sforallvari x{}}}
   \tight+\underline{\app{\wforallvari g{}}{\sforallvari x{}}}}}}
   -\framebox{\math{\inpit{\underline{\wforallvari y f}\tight+
                           \underline{\wforallvari y g}}}}\,\,|
 \nottight{\nottight{\nottight<}}
 \lfloor\wforallvari\varepsilon{}\rfloor
}

\halftop\noindent
Now the essential idea of the whole proof is to 
apply the lemma (5) of \figuref{figure global abbreviations} via \\\math{\{
\wforallvari z 0\mapsto\app{\wforallvari f{}}{\sforallvari x{}}\comma
\wforallvari z 1\mapsto\app{\wforallvari g{}}{\sforallvari x{}}\comma
\wforallvari z 2\mapsto\wforallvari y f\comma
\wforallvari z 3\mapsto\wforallvari y g
\}}, by which we get:
\par\noindent\math{(1^5.3.1^2.2.1.5.1)}:
\ \ \math{
  \framebox{\framebox{\math{t\nleq
  |\app{\wforallvari f{}}{\sforallvari x{}}\tight-\wforallvari y f|
  +
  |\app{\wforallvari g{}}{\sforallvari x{}}\tight-\wforallvari y g|}}}
    \comma
  \framebox{\math{t<\wforallvari\varepsilon{}}}
    \comma
}\\\mbox{}\hfill\math{
        |\app{\wforallvari f{}}{\sforallvari x{}}\tight-\wforallvari y f|
        \nless\existsvari\varepsilon f
  \comma
        |\app{\wforallvari g{}}{\sforallvari x{}}\tight-\wforallvari y g|
        \nless\existsvari\varepsilon g
  \comma
  \ldots}

\subsection{Automatic Clean-Up}\label{section clean up}
The rest of the proof is perfectly within the scope of automatic proof search
today. When we apply the other transitivity lemma (6) 
of \figuref{figure global abbreviations}
to \math{(1^5.3.1^2.2.1.5.1)} as indicated by the single and double boxes
in the goal and the lemma, via 
\bigmaths{\{\ \ 
  \wforallvari z 4\mapsto t 
 \comma
  \wforallvari z 6\mapsto\wforallvari\varepsilon{}
 \comma
  \wforallvari z 5\mapsto
  |\app{\wforallvari f{}}{\sforallvari x{}}\tight-\wforallvari y f|
  +
  |\app{\wforallvari g{}}{\sforallvari x{}}\tight-\wforallvari y g|
\ \ \}},
we get:
\par\halftop\noindent\math{(1^5.3.1^2.2.1.5.1^2)}:
\ \ \math{
    |\app{\wforallvari f{}}{\sforallvari x{}}\tight-\wforallvari y f|
    +
    |\app{\wforallvari g{}}{\sforallvari x{}}\tight-\wforallvari y g|
    <
    \wforallvari\varepsilon{}    
  \comma
}\nopagebreak\\\mbox{}\hfill\math{
\framebox{\math{
        |\app{\wforallvari f{}}{\sforallvari x{}}\tight-\wforallvari y f|
        \nless\existsvari\varepsilon f}}
  \comma
\framebox{\framebox{\math{
        |\app{\wforallvari g{}}{\sforallvari x{}}\tight-\wforallvari y g|
        \nless\existsvari\varepsilon g}}}
  \comma\ldots}

\begin{sloppypar}\noindent In \cite{colored-rippling-lim-+} even the step 
from 
\math{(1^5.3.1^2.2.1.5)}
to
\math{(1^5.3.1^2.2.1.5.1^2)}
is automated 
with the wave-front annotation of 
\bigmath{t<\wforallvari\varepsilon{}}
as given in \nolinebreak\sectref{section immediate focus}
\ (which is generated by the givens of
 \ \mbox{\maths
{|\app{\wforallvari f{}}{\sforallvari x{}}\tight-\wforallvari y f|
        <\existsvari\varepsilon f}{}} \ 
and
\bigmaths{|\app{\wforallvari g{}}{\sforallvari x{}}\tight-\wforallvari y g|
        <\existsvari\varepsilon g}{}
in the context of \bigmaths{t<\wforallvari\varepsilon{}}{} 
in \math{(1^5.3.1^2.2.1.5)}), \ 
provided that the following lemmas (annotated as wave-rules) are 
in the rippling system:\end{sloppypar}
\halftop\noindent\LINEmath{
   \framebox{\math{\inpit{\underline{\wforallvari z 0}\tight+
                          \underline{\wforallvari z 1}}}}
   -
   \framebox{\math{\inpit{\underline{\wforallvari z 2}\tight+
                          \underline{\wforallvari z 3}}}}
  \nottight{\nottight=}
  \framebox{\math{\underline{\inpit{\wforallvari z 0\tight-\wforallvari z 2}}
  +
  \underline{\inpit{\wforallvari z 1\tight-\wforallvari z 3}}}}
}
\par\noindent\LINEmath{
  |~\framebox{\math{\underline{\wforallvari z 4}+
                    \underline{\wforallvari z 5}}}~|<
  \wforallvari z 6
  \comma~~
  \framebox{\math{
  \underline{|\wforallvari z 4|}
  +
  \underline{|\wforallvari z 5|}}}
  \nless
  \wforallvari z 6
}

\halftop\noindent
Applying lemma \nlbmath{(7)} of \figuref{figure global abbreviations}
(monotonicity of \math+) in the obvious way, we get:
\par\halftop\noindent\math{(1^5.3.1^2.2.1.5.1^3)}:
\ \ \math{\begin{array}[t]{l}    
    |\app{\wforallvari f{}}{\sforallvari x{}}\tight-\wforallvari y f|
    +
    |\app{\wforallvari g{}}{\sforallvari x{}}\tight-\wforallvari y g|
    \nless
    \existsvari\varepsilon f
    +
    \existsvari\varepsilon g
  \comma
\\|\app{\wforallvari f{}}{\sforallvari x{}}\tight-\wforallvari y f|
    +
    |\app{\wforallvari g{}}{\sforallvari x{}}\tight-\wforallvari y g|
    <
    \wforallvari\varepsilon{}    
  \comma\ldots
\\\end{array}}

\begin{sloppypar}\noindent
The \math R-substitution \bigmaths{\{
    \existsvari\varepsilon f\tight\mapsto{\wforallvari\varepsilon{}\over 2}
  \comma
    \existsvari\varepsilon g\tight\mapsto{\wforallvari\varepsilon{}\over 2}
\}}{}
closes the remaining open branches of \math{(1^5.3.1^2.2.1.5.1^3)}
and \math{(1^5.[\mbox{1--2}])}
with the lemmas \nlbmath{(3),(8)} and \nlbmath{(9)}, respectively. 
The final \vc\ is acyclic indeed.
Its graph is depicted in 
\figuref{figure variable-condition final} below.  
The whole proof tree with a minor permutation of the critical 
\math\beta-step is depicted in \figurefsix\ in 
\sectref{section defining}.\vfill\pagebreak\end{sloppypar}

\figurefinalvc h

\yestop\yestop\yestop\yestop\figuredeltaminusinstead h

\yestop\yestop\section{Discussion}\label
{section discussion}\yestop\noindent
Now that the \nonpermutability\ 
of \protect\math\beta\ 
at \protect\nlbmath{(1^5.3.1)} and 
\protect\deltaplus\ at \protect\nlbmath{(1^5.3.1.2)}
(\cfnlb\ \figuref{figure failure}) as well as the \nonpermutability\ 
of \protect\math\beta\ 
at \nlbmath{(1^5.3.1^2)} and 
\protect\math{\beta} at \nlbmath{(1^5.3.1^2.2)}
(\cfnlb\ \figuref{figure vision})
have become practically evident 
by the proof of {\nolinebreak$($$\lim$$+$$)$} in \sectref{section proof}, 
\mbox{we may ask:} \ 
{\em Why did the co-lecturer not believe in what he saw?}

He knew that the only problem with the sequencing of 
\mbox{\math\beta-steps} 
that occurs either with the \deltaminus-rules 
or else with the \deltaplusplus-rules \nlbcite{deltaplusplus} 
is that a bad choice makes the proofs suffer from
the repetition of common sub-proofs,
which is an optimization problem 
not subsumed under the notion of \nonpermutability, 
\cfnlb\ \sectref{section introduction}. \ 

Thus, we have to make it even clearer
why the \deltaplus-rules are so much in conflict with the \math\beta-steps.
\vfill\pagebreak

\yestop\subsection
{\NonPermutability\ of \protect\math\beta\ and \protect\math\beta\ is only a 
 Secondary Problem}%\figureaftersigma b%
Notice that the \nonpermutability\ of \math\beta\
and \deltaplus\ is the primary problem and the only one we have to explain.
It causes the \nonpermutability\ of \math\beta\ and \math\beta\ we have 
seen in \figuref{figure vision} as a secondary problem:
Indeed, 
the \nth 2 \math\beta-step
in \figuref{figure vision} must come before the \nth 1 \math\beta-step 
simply because the \nth 2 \math\beta-step 
generates the principal \mbox{\math\delta-formula} of the 
\app{\delta_0^+}{\sforallvari x{}}-step resulting in the rightmost leaf,
and this \app{\delta_0^+}{\sforallvari x{}}-step 
must come before the \nth 1 \math\beta-step;
namely for the leftmost leaf's first formula \nlbmath
{\protect\existsvari x f\protect\boldunequal\protect\wforallvari x 0}
to be of any use in the proof.
This means that 
\par\halftop\noindent\LINEmaths
{\nth 2\math\beta 
 \nottight{\nottight{\nottight{<_{\mbox{\rm\scriptsize superformula}}}}}
 \app{\delta_0^+}{\sforallvari x{}} 
 \nottight{\nottight{\nottight{<_
 {\mbox{\rm\scriptsize\math\beta-\deltaplus-\nonpermutability}}}}}
 \nth 1\math\beta
}{}\par\halftop\noindent
causes the \nonpermutability\ of \nth 1\math\beta\ and \nth 2\math\beta\ 
by transitivity.

\subsection{\deltaminus\ instead of \deltaplus}\label{section minus}
Let us see how the proof of {\nolinebreak$($$\lim$$+$$)$} would look like
with the \mbox{\deltaminus-rules} as the only \mbox{\math\delta-rules} 
available.
Roughly speaking, in the proof of \sectref{section proof},
we have to replace each \sfuv\ \nlbmath{\sforallvari v n}
with a \wfuv\ \nlbmath{\wforallvari v n} and check
how the \vc\ changes: \ 
\app{\delta_0^-}{\wforallvari\delta f} \nolinebreak
and \nlbmath{\app{\delta_0^-}{\wforallvari\delta g}} 
applied to 
\nlbmath{(1^5.3)} of \sectref{section expanding the proof} and 
\nlbmath{(1^5.3.1.2)} of \sectref{section bad turn}
(\cfnlb\ \figuref{figure failure}) add
\bigmaths{\{\existsvari\varepsilon f,\existsvari\varepsilon g,
\existsvari\delta{}\}\times\{\wforallvari\delta f\}}{} 
and \bigmaths{\{\existsvari\varepsilon f,\existsvari\varepsilon g,
\existsvari\delta{}\}\times\{\wforallvari\delta g\}}{}
to the initially empty \vc\ \nlbmath R,
respectively. \ 
\app{\delta_0^-}{\wforallvari x{}} \nolinebreak applied roughly at
\nlbmath{(1^5.3.1.2.1)}
adds \bigmaths{\{\existsvari\varepsilon f,\existsvari\varepsilon g,
\existsvari\delta{}\}\times\{\wforallvari x{}\}}{} 
later.

\yestop\noindent 
Thus, \ after applying 
\par\halftop\noindent\LINEmath{
  \sigma^-
  :=
  \{
  \existsvari x f\tight\mapsto\wforallvari x{}\comma
  \existsvari x g\tight\mapsto\wforallvari x{}\comma
  \existsvari\delta{}\tight\mapsto
    \min(\wforallvari\delta f,\wforallvari\delta g)
  \}
}\par\halftop\noindent
the \mbox{\math{\sigma^-}-updated} \vc\ is extended by 
\par\halftop\noindent\LINEmaths{\{ 
\pair{\wforallvari x{}}{\existsvari x f},
\pair{\wforallvari x{}}{\existsvari x g},
\pair{\wforallvari\delta f}{\existsvari\delta{}},
\pair{\wforallvari\delta g}{\existsvari\delta{}}\}}{}\par\halftop\noindent
and looks as in \figuref{figure delta minus} above.
Compared to the graph of \figuref{figure variable-condition one},
it is small but cyclic: 
Among others, the two curved edges
at the very bottom are new and cause the cycles. Thus, 
\math{\sigma^-}\nolinebreak\ 
is no \math R-substitution at all and cannot be applied.

Therefore, in our example proof of \sectref{section proof} as depicted in 
\figuref{figure failure}, we have to move
the \math\gamma-step applied to \math{(1^3)} down below
\nlbmath{(1^5.3.1.2.1)}. \ 
Note that we cannot move it deeper because it has to preceed 
the step \nlbmath{\app{\delta_0^-}{\wforallvari x{}}}: Indeed,
the principal formula of this \deltaminus-step
is a subformula of the side formula of the \math\gamma-step. 
{A fortiori}, this movement of the \math\gamma-step applied to 
\nlbmath{(1^3)} forces 
the problematic \math\beta-step at \nlbmath{(1^5.3.1)} to be moved below 
\nlbmath{(1^5.3.1.2.1)}, \ 
too; \ simply because its principal \math\beta-formula is 
the side formula of the \math\gamma-step. \ 

Indeed, if we replace the \deltaplus-rules with \deltaminus-rules,
the \nonpermutability\ of the \math\beta- and the 
\deltaplus-steps is hidden behind the well-known \nonpermutability\ 
of the \math\gamma- and the \deltaminus-steps,
\cfnlb\ \sectref{section introduction}. \ 
Only when the latter \nonpermutability\ is removed by replacing
the \deltaminus-rules with \deltaplus-rules, 
the former becomes visible.
\vfill\pagebreak

\subsection{\SFUV s can Escape their Quantifiers' Scopes}\label
{section escape their quantifiers scopes}\label
{section where all the different deltas are referenced}
The \nonpermutability\ of the \math\beta- and \deltaplus-steps is closely
related to the following strange aspect of the \deltaplus-rules, 
which they share with 
the \mbox{\deltaplusplus-rules} \nlbcite{deltaplusplus},
the \mbox{\deltastar-rules} \nlbcite{baazdelta}, and
the 
\mbox{\deltastarstar-rules \nlbcite{deltasuperdot}}, \ 
but not with the 
\math{\delta^\varepsilon}-rules \cite{ahrendtgiese} and the \deltaminus-rules. \
While soundness % and safeness 
of both the \deltaminus- and \deltaplus-rules
and preservation of solutions of the \deltaminus-rules are immediate, 
the preservation of solutions 
of the \deltaplus-rules requires the restriction of the values of 
the \sfuv s by \cc s \cite[\littheoref{2.49}]{wirthcardinal}. \ 
Although there is no space here for introducing the semantics 
of the several kinds of free variables of \cite{wirthcardinal},
the reader may grasp the idea of the following example, 
namely that a solution for \nlbmath{\existsvari x{}}
that makes the lower sequent true, may make the upper sequent false:

\yestop\begin{example}[{Reduction \& Liberalized \math\delta, 
\cite[\litexamref{2.29}]{wirthcardinal}}]
\label{example eta}\newcommand\outdent{\hskip 6em\mbox{}}%
\nopagebreak\mbox{}\\In \cite[\litexamref{2.8}]{wirthcardinal},
a \deltaplus-step reduces
\hfill\math{
      \forall y\stopq\neg\Pppp y
  \comma~~ 
      \Pppp{\existsvari x{}}
  \comma
      \ldots
}\outdent\nopagebreak\\
to
\hfill\math{
      \neg\Pppp{\sforallvari y{}}
  \comma~~
      \Pppp{\existsvari x{}}
  \comma
      \ldots
}\outdent\\\noindent
with the empty \vc\ \math{R:=\emptyset}.\yestop\end{example} 
Let us first argue semantically:
The lower sequent is \pair e\salgebra-valid for the 
\strongexRval\ \nlbmath e given by
\par\noindent\LINEmaths{
  \app{\app e{\existsvari x{}}}\delta
  :=
  \app\delta{\sforallvari y{}}
},\par\halftop\noindent
which sets the value of \existsvari x{} to the value of \sforallvari y{}. \ 
The upper sequent, however, is not \pair e\salgebra-valid 
when \app{\Ppsymbol^\salgebra}a is \TRUEpp\ 
and  \app{\Ppsymbol^\salgebra}b is \FALSEpp\
for some \math a, \math b from the universe of the structure 
\nolinebreak\salgebra. \ 
To \nolinebreak see this, take some valuation \nlbmath\delta\ 
with \math{\app\delta{\sforallvari y{}}:=b}. \ 
Then \existsvari x{} and \sforallvari y{} both evaluate to \nlbmath b, 
the lower sequent to \bigmaths{\TRUEpp\comma\FALSEpp}, and
the upper sequent to \bigmaths{\FALSEpp\comma\FALSEpp}.

No matter whether this semantical argumentation can become clear here, 
the following syntactical variant will \nolinebreak do similarly well: 
After applying the \mbox{\math R-substitution}
\par\halftop\noindent\LINEmaths
{\mu^+:=\{\existsvari x{}\tight\mapsto\sforallvari y{}\}},\par\halftop\noindent
the lower sequent is a tautology, whereas the upper sequent is not.

\yestop\noindent
This cannot happen with the \deltaminus-rules: \ 
Their application instead of the \deltaplus-rules adds 
\math{\{\pair{\existsvari x{}}{\wforallvari y{}}\}} to the \vc, 
thereby blocking 
\par\halftop\noindent\LINEmaths
{\mu^-:=\{\existsvari x{}\tight\mapsto\wforallvari y{}\}},\par\halftop\noindent
simply because \nlbmath {\mu^-}
is no \math{\{\pair{\existsvari x{}}{\wforallvari y{}}\}}-sub\-sti\-tu\-tion,
\cfnlb\ \defiref{definition ex r sub}. \ 

From a semantical point of view, however, 
the \nlbmath e displayed above is no \strongexRval\ 
for the extended \vc\ anymore.

\yestop\noindent
Roughly speaking, via \nlbmath{\mu^+}, the \deltaplus-variable 
\nlbmath{\sforallvari y{}} escapes the scope of the quantifier
\nlbmath{\forall y} on the bound variable \nlbmath{\boundvari y{}} which was 
eliminated by the introduction of \nlbmath{\sforallvari y{}}.
At least with matrix calculi and indexed formulas trees 
\nlbcite{sergediss,wallen}, \ 
this ``escaping'' 
is a natural way to talk about this strange
liberality of the \mbox{\deltaplus-rule}.
And it also happens in \figuref{figure failure} of the proof of 
{\nolinebreak$($$\lim$$+$$)$}: \ 
Taking the tree of \figuref{figure failure}
to be an indexed formula tree, roughly speaking,
the quantifier for \nlbmath{\sforallvari\delta g} is
situated at the term position \nlbmath{(1^5.3.1.2)}, but, via \nlbmath\sigma,
it escapes to term position \nlbmath{(1^5.3.1.1)}.

\yestop\subsection{\deltaplusplus\ instead of \deltaplus}\label
{section deltaplusplus}Let us see how the proof of 
{\nolinebreak$($$\lim$$+$$)$} would look like
with the \mbox{\deltaplusplus-rules} \cite{deltaplusplus}
as the only \mbox{\math\delta-rules} available.
This does not change anything in the proof as given in \sectref{section proof},
but allows us to use 
the identical \sfuv\ %\sforallvari\delta f and 
\sforallvari\delta g again when repeating the 
\math{\delta}-step which introduced it.
Thus, starting from \nlbmath{(1^5.3.1.1)} of \sectref{section bad turn}, 
we can repeat some of the steps
done in proof of \nlbmath{(1^5.3.1.2)}, namely
\ ``~\app{\delta_0^{+}}{\sforallvari\delta g}, 
\math{\footroom\alpha_0^2}~'' \  of
\figuref{figure failure}, but now as 
\ ``~\app{\delta_0^{+^+}}{\sforallvari\delta g}, 
\math{\alpha_0^2}~\closequotefullstop 
Note that the \deltaplus-rules would allow 
\app{\delta_0^+}{\sforallvari\delta G} only, with new
\sforallvari\delta G. \ 
The resulting sequent is
\par\yestop\noindent\math{{(1^5.3.1.1.1)}}:
\mbox{} \ \math{0\tight<\min(\sforallvari\delta f,\sforallvari\delta g)
\comma\Omega}\par\yestop\noindent
It is like \maths{(1^5.3.1.2.1)}{}
of \sectref{section bad turn}, but with the \math{\beta_2}-side formula
of the critical \math\beta-step replaced with the \math{\beta_1}-side formula
\bigmaths{0\tight<\min(\sforallvari\delta f,\sforallvari\delta g)}.
This formula admits to close this branch with the formulas
\bigmaths{0\tight\nless\sforallvari\delta f}{}
and
\bigmaths{0\tight\nless\sforallvari\delta g}{}
(as listed in \nlbmath\Omega\ of \figuref{figure global abbreviations}),
applying lemma \nlbmath{(4)} of \figuref{figure global abbreviations}
as at the end of \sectref{section backtracking}.

Notice that this proof with the \deltaplusplus-rules does not have a higher
number of \mbox{\math\gamma-steps} than the proof attempt failing
in \sectref{section total failure}. \ 
Also the maximum number of \math\delta-steps per
formula and\emph{per path} is still \nlbmath 1. \ 
Nevertheless, 
the multiple expansion of the same \math\delta-formula in different
paths is somehow counter-intuitive and nothing a working mathematician would
expect. 
In indexed formula trees based on the \deltaplusplus-rules,
all \math\delta-formulas are treated only once. 
This again means that these matrix
versions are more human-oriented than the tableau or sequent versions.
\vfill\pagebreak

\section{Proof of the \NonPermutability\ of \math\beta\ and \deltaplus}\label
{section meta-proof}

As we have seen in \sectref{section minus}, 
the \nonpermutable\ \math\beta-step necessarily follows a 
\math\gamma-step that would be \nonpermutable\ without the \mbox{liberalization}
\mbox{from \deltaminus\ to \deltaplus}.
It follows indeed\emph{necessarily},
because the principal formula of the \math\beta-step is the side formula
of the \math\gamma-step. 
Although \begin{itemize}\item the \mbox{\math\gamma-step 
\nlbmath{\app{\gamma_0}{\min(\sforallvari\delta f,\sforallvari\delta g)}}}
is permutable with the liberalized
\deltaplus-step \nlbmath{\app{\delta_0^+}{\sforallvari\delta g}},
\item\sloppy the \mbox{\math\gamma-step 
\nlbmath{\app{\gamma_0}{\min(\wforallvari\delta f,\wforallvari\delta g)}}},
however, 
is \nonpermutable\ with the \mbox
{\deltaminus-step \math{\app{\delta_0^-}{\wforallvari\delta g}},}\end{itemize}
and even with the liberalization\begin{itemize}\item
the \math\beta-step is still \nonpermutable\ with the 
\mbox{\deltaplus-step \nlbmath{\app{\delta_0^+}{\sforallvari\delta g}}}.\end
{itemize}
As the principal formula of the \math\beta-step can be regenerated
by a second expansion of the principal formula of the \math\gamma-step,
we cannot prove the \nonpermutability\ unless we restrict the 
\mbox{\math\gamma-multiplicity}. \ 
But, according to the description of the notion of 
\nonpermutability\ in \sectref{section introduction},
we may indeed restrict the 
\math\gamma-multiplicity, in which case the crucial step, 
namely \lemmref{lemma no completion},
admits the following semantical proof.

\subsection{Proof of \lemmref{lemma no completion}
at the end of \sectref{section total failure}}\label{section proof of lemma}
Let us remove the three 
\mbox{\math\gamma-formulas} which form the sequent \nlbmath\Gamma\ 
(\cfnlb\ \figuref{figure global abbreviations})
from the sequents \math{(1^5.1)}, \math{(1^5.2)} 
(\cfnlb\ \sectref{section expanding the proof}), and 
\nlbmath{(1^5.3.1.1)} (\cfnlb\ \sectref{section bad turn}).
As these \math\gamma-formulas were already once expanded at \nlbmath{(1^3)}
and \nlbmath{(1^4)} (\cfnlb\ \figuref{figure failure}),
this removal represents a restriction of the \math\gamma-multiplicity
of the removed \math\gamma-formulas to \nlbmath 1,
and results in the following sequents (after some reordering):
\par\yestop\noindent\math{{(1^5.1\tightsetminus\Gamma+)}}:
\ \ \ \ \math{0\tight<\existsvari\varepsilon f\comma
0\tight\nless\wforallvari\varepsilon{}\comma
}\\\mbox{}\hfill\math{
    \neg
    \inparentheses{
    0\tight<\existsvari\varepsilon g
    \implies
    \exists\delta_g\tight>0\stopq 
    \forall x_g\boldunequal\wforallvari x 0\stopq 
    \inparenthesesoplist{
        |\app{\wforallvari g{}}{x_g}\tight-\wforallvari y g|<\existsvari\varepsilon g
      \oplistantiimplies
        |x_g\tight-\wforallvari x 0|<\delta_g
    }}
  \comma
}\\\mbox{}\hfill\math{
    0\tight<\min(\sforallvari\delta f,\sforallvari\delta g)    
    \und
    \forall x\boldunequal\wforallvari x 0\stopq 
    \inparenthesesoplist{
        |\inpit{\app{\wforallvari f{}}{x}\tight+\app{\wforallvari g{}}{x}}
         \tight-\inpit{\wforallvari y f\tight+\wforallvari y g}|
        <\wforallvari\varepsilon{}
      \oplistantiimplies
        |x\tight-\wforallvari x 0|
        <\min(\sforallvari\delta f,\sforallvari\delta g)
    }
}
\par\yestop\noindent\math{{(1^5.2\tightsetminus\Gamma+)}}:
\ \ \ \ \math{0\tight<\existsvari\varepsilon g\comma
0\tight\nless\wforallvari\varepsilon{}\comma
}\\\mbox{}\hfill\math{
    \neg\exists\delta_f\tight>0\stopq 
    \forall x_f\boldunequal\wforallvari x 0\stopq 
    \inparenthesesoplist{
        |\app{\wforallvari f{}}{x_f}\tight-\wforallvari y f|<\existsvari\varepsilon f
      \oplistantiimplies
        |x_f\tight-\wforallvari x 0|<\delta_f
    }\comma
}\\\mbox{}\hfill\math{
    0\tight<\min(\sforallvari\delta f,\sforallvari\delta g)    
    \und
    \forall x\boldunequal\wforallvari x 0\stopq 
    \inparenthesesoplist{
        |\inpit{\app{\wforallvari f{}}{x}\tight+\app{\wforallvari g{}}{x}}
         \tight-\inpit{\wforallvari y f\tight+\wforallvari y g}|
        <\wforallvari\varepsilon{}
      \oplistantiimplies
        |x\tight-\wforallvari x 0|
        <\min(\sforallvari\delta f,\sforallvari\delta g)
    }
}
\par\yestop\noindent\math{(1^5.3.1.1\tightsetminus\Gamma+)}: 
\math{\ \ \ \ 
0\tight<\min(\sforallvari\delta f,\sforallvari\delta g)
\comma
0\tight\nless\wforallvari\varepsilon{}
\comma 
}\nopagebreak\\\mbox{}\hfill\math{
\oplistvarilesssideformuladeltaplusnegexistsdelta f\sforallvari\comma
}\nopagebreak\\\mbox{}\hfill\math{
\oplistprincipalformuladeltaplusnegexistsdelta g
}\par\yestop\noindent
The related \vc\ \nlbmath R \ 
is shown in \figuref{figure variable-condition one} (without the dotted edges)
and the current \math R-\cc\ \nlbmath C is given as
\par\noindent\mbox{}\hfill\math{\displayset{\!\!
  {\sforallvari x{}}\mapsto
  {\neg\inparentheses{
    {\sforallvari x{}}\boldunequal\wforallvari x 0
    \implies
    \inparenthesesoplist{
        |\inpit{\app{\wforallvari f{}}{\sforallvari x{}}
           \tight+\app{\wforallvari g{}}{\sforallvari x{}}}
         \tight-\inpit{\wforallvari y f\tight+\wforallvari y g}|
        <\wforallvari\varepsilon{}
      \oplistantiimplies
         |{\sforallvari x{}}\tight-\wforallvari x 0|
        <\min(\sforallvari\delta f,\sforallvari\delta g)
    }\!\!\!\!}},\!\!\!\!\\\!\!
   {\sforallvari\delta f}\mapsto\inparentheses
   {0\tight<\sforallvari\delta f\und
    \forall x_f\boldunequal\wforallvari x 0\stopq 
    \inparenthesesoplist{
        |\app{\wforallvari f{}}{x_f}\tight-\wforallvari y f|<\existsvari\varepsilon f
      \oplistantiimplies
        |x_f\tight-\wforallvari x 0|<\sforallvari\delta f
    \!\!}\!\!\!\!},\!\!\!\!\\\!\!
   {\sforallvari\delta g}\mapsto\inparentheses
   {0\tight<\sforallvari\delta g\und
    \forall x_g\boldunequal\wforallvari x 0\stopq 
    \inparenthesesoplist{
        |\app{\wforallvari g{}}{x_g}\tight-\wforallvari y g|<\existsvari\varepsilon g
      \oplistantiimplies
        |x_g\tight-\wforallvari x 0|<\sforallvari\delta g
}\!\!\!\!}}}\par\yestop\yestop\noindent
It now suffices to show that there is no proof of 
\math{(1^5.1\tightsetminus\Gamma+)},
\math{(1^5.2\tightsetminus\Gamma+)}, and
\nlbmath{(1^5.3.1.1\tightsetminus\Gamma+)}
with the \math\deltaminus- and \deltaplus-rules as the only
\math\delta-rules available.
 
\yestop\noindent
We do this with a trivial transformation given by the substitution 
\par\yestop\noindent\LINEmaths{\nu:=\{
\sforallvari\delta f\tight\mapsto\wforallvari\delta f
\comma
\sforallvari\delta g\tight\mapsto\wforallvari\delta g
\}}{}\par\yestop\noindent
of an assumed proof of 
\math{(1^5.1\tightsetminus\Gamma+)},
\math{(1^5.2\tightsetminus\Gamma+)}, and
\nlbmath{(1^5.3.1.1\tightsetminus\Gamma+)}
on the one hand,
and with a deviation over invalidity
and soundness on the other hand, 
as follows:

\yestop\noindent
Instantiating the sequents
\math{(1^5.1\tightsetminus\Gamma+)},
\math{(1^5.2\tightsetminus\Gamma+)}, and
\nlbmath{(1^5.3.1.1\tightsetminus\Gamma+)}
by \nlbmath\nu\ we get the sequents
\par\yestop\noindent\math{{(1^5.1\tightsetminus\Gamma-)}}:
\ \math{0\tight<\existsvari\varepsilon f\comma
0\tight\nless\wforallvari\varepsilon{}\comma
}\\\mbox{}\hfill\math{
    \neg
    \inparentheses{
    0\tight<\existsvari\varepsilon g
    \implies
    \exists\delta_g\tight>0\stopq 
    \forall x_g\boldunequal\wforallvari x 0\stopq 
    \inparenthesesoplist{
        |\app{\wforallvari g{}}{x_g}\tight-\wforallvari y g|<\existsvari\varepsilon g
      \oplistantiimplies
        |x_g\tight-\wforallvari x 0|<\delta_g
    }}
  \comma
}\\\mbox{}\hfill\math{
    0\tight<\min(\wforallvari\delta f,\wforallvari\delta g)    
    \und
    \forall x\boldunequal\wforallvari x 0\stopq 
    \inparenthesesoplist{
        |\inpit{\app{\wforallvari f{}}{x}\tight+\app{\wforallvari g{}}{x}}
         \tight-\inpit{\wforallvari y f\tight+\wforallvari y g}|
        <\wforallvari\varepsilon{}
      \oplistantiimplies
        |x\tight-\wforallvari x 0|
        <\min(\wforallvari\delta f,\wforallvari\delta g)
    }
}
\par\yestop\noindent\math{{(1^5.2\tightsetminus\Gamma-)}}:
\ \math{0\tight<\existsvari\varepsilon g\comma
0\tight\nless\wforallvari\varepsilon{}\comma
    \neg\exists\delta_f\tight>0\stopq 
    \forall x_f\boldunequal\wforallvari x 0\stopq 
    \inparenthesesoplist{
        |\app{\wforallvari f{}}{x_f}\tight-\wforallvari y f|<\existsvari\varepsilon f
      \oplistantiimplies
        |x_f\tight-\wforallvari x 0|<\delta_f
    }\comma
}\\\mbox{}\hfill\math{
    0\tight<\min(\wforallvari\delta f,\wforallvari\delta g)    
    \und
    \forall x\boldunequal\wforallvari x 0\stopq 
    \inparenthesesoplist{
        |\inpit{\app{\wforallvari f{}}{x}\tight+\app{\wforallvari g{}}{x}}
         \tight-\inpit{\wforallvari y f\tight+\wforallvari y g}|
        <\wforallvari\varepsilon{}
      \oplistantiimplies
        |x\tight-\wforallvari x 0|
        <\min(\wforallvari\delta f,\wforallvari\delta g)
    }
}
\par\yestop\noindent\math{(1^5.3.1.1\tightsetminus\Gamma-)}: 
\hfill\math{
0\tight<\min(\wforallvari\delta f,\wforallvari\delta g)\comma
0\tight\nless\wforallvari\varepsilon{}\comma 
}\nopagebreak\\\mbox{}\hfill\math{
\varilesssideformuladeltaplusnegexistsdelta f\wforallvari\comma
}\nopagebreak\\\mbox{}\hfill\math{
\oplistprincipalformuladeltaplusnegexistsdelta g}\par\yestop\noindent
The conjunction of these sequents is invalid according to the standard 
seman\-tics for parameters
as well as the semantics of \nlbcite{wirthcardinal}.
This can be seen by
\par\yestop\noindent\LINEmaths{\{\ 
\wforallvari\delta f     \tight\mapsto 1,\hfill
\wforallvari\delta g     \tight\mapsto 0,\hfill
\wforallvari\varepsilon{}\tight\mapsto 1,\hfill
\wforallvari x 0         \tight\mapsto 0,\hfill
\wforallvari y f         \tight\mapsto 0,\hfill
\wforallvari y g         \tight\mapsto 0,\hfill
\wforallvari f{}         \mapsto\lambda x.0,\hfill
\wforallvari g{}         \mapsto\lambda x.0\ \}}.\par\yestop\noindent
Indeed, if we instantiate 
\math{(1^5.1\tightsetminus\Gamma-)},
\math{(1^5.2\tightsetminus\Gamma-)}, and
\nlbmath{(1^5.3.1.1\tightsetminus\Gamma-)}
with this substitution
and then \mbox{\math{\lambda\beta}-normalize} and simplify 
these sequents by equivalence transformations in the 
model of the real numbers \nlbmath\RE, we get the three sequents
\par\yestop\noindent
\LINEmath{0\tight<\existsvari\varepsilon f\comma\falsepp\comma
    \neg
    \inparentheses{
    0\tight<\existsvari\varepsilon g
    \implies
    \inparenthesesoplist{
        0\tight<\existsvari\varepsilon g
      \oplistantiimplies
        \forall\delta_g\tight>0\stopq 
        \exists x_g\boldunequal 0\stopq 
          |x_g|\tight<\delta_g
    }}\comma\falsepp
}
\par\yestop\noindent
\LINEmath{0\tight<\existsvari\varepsilon g\comma\falsepp\comma
    \neg
    \inparenthesesoplist{
        0\tight<\existsvari\varepsilon f
      \oplistantiimplies
        \forall\delta_f\tight>0\stopq 
        \exists x_f\boldunequal 0\stopq 
          |x_f|\tight<\delta_f
    }\comma\falsepp
}
\par\yestop\noindent\LINEmath{\falsepp\comma\falsepp\comma
\neg\inpit{
  0\tight<\existsvari\varepsilon f\antiimplies
  \exists\boundvari x f\boldunequal 0\stopq |\boundvari x f|\tight<1}\comma
\neg\inparenthesesoplist{
        0\tight<\existsvari\varepsilon g
      \oplistantiimplies
        \forall\boundvari\delta g\tight>0\stopq
          \exists\boundvari x g\boldunequal 0\stopq
          |\boundvari x g|\tight<\boundvari\delta g}
}\par\yestop\noindent
Further equivalence transformation in \RE\ results in the three 
contradictory sequents
\par\yestop\noindent\LINEmath{
   0\tight<\existsvari\varepsilon f
}
\par\yestop\noindent\LINEmath{
   0\tight<\existsvari\varepsilon g\comma
   0\tight\nless\existsvari\varepsilon f
}
\par\yestop\noindent\LINEmath{
   0\tight\nless\existsvari\varepsilon f\comma
   0\tight\nless\existsvari\varepsilon g
}\par\yestop\noindent
Thus, as our calculus is sound, it cannot prove 
\math{(1^5.1\tightsetminus\Gamma-)},
\math{(1^5.2\tightsetminus\Gamma-)}, and
\nlbmath{(1^5.3.1.1\tightsetminus\Gamma-)} in parallel.

\yestop\yestop\noindent
As the \deltaplus-rules treat free \deltaminus- and \sfuv s alike,
and as the \mbox{\deltaminus-rules} generate a smaller \vc\ for
free \deltaminus- instead of \sfuv s in the principal sequents
(\cf\ \VARsomesall{\ldots} in \figuref{figure reductive rules}),
a proof of 
\math{(1^5.1\tightsetminus\Gamma+)},
\math{(1^5.2\tightsetminus\Gamma+)}, and
\nlbmath{(1^5.3.1.1\tightsetminus\Gamma+)}
would immediately translate into a proof of 
\math{(1^5.1\tightsetminus\Gamma-)},
\math{(1^5.2\tightsetminus\Gamma-)}, and
\nlbmath{(1^5.3.1.1\tightsetminus\Gamma-)}
with unchanged inference rules,
just by \nolinebreak application of the substitution \nlbmath\nu.

\yestop\yestop\noindent Thus, we conclude that there is no proof of 
\math{(1^5.1\tightsetminus\Gamma+)},
\math{(1^5.2\tightsetminus\Gamma+)}, and
\math{(1^5.3.1.1\tightsetminus\Gamma+)}. \qedabbrev

\yestop\yestop\yestop\noindent
Note that the above trivial proof transformation does not 
result in a sound proof if we replace the 
\deltaplus-rules with the \deltaplusplus-rules:
Indeed, the \deltaplusplus-rules may re-use 
\nolinebreak\sforallvari\delta g, but not
\nolinebreak\wforallvari\delta g.
\vfill\pagebreak

\yestop\yestop
\subsection{Defining Permutability}\label{section defining}

\noindent
A reader with a good mathematical intuition can and should 
directly consider the \nonpermutability\ of \math\beta- and \deltaplus-steps
as a corollary of \lemmref{lemma no completion} proved above.
A formalist, however, may well require 
some rigorous definition of permutability.
There were good reasons not to present a formal definition of permutability 
earlier in this \daspaper:
\begin{enumerate}
\halftop\item
The logically weakest reasonable definitions of permutability I can think of,
still result in the \nonpermutability\ we want to show. \ 
Indeed, we may choose any definition of permutability that 
contradicts \lemmref{lemma no completion}. \ 
For instance, as it strengthens our 
\nonpermutability\ result, we should (and will)
use a notion that is weaker than the following standard one:
Two inference steps \nlbmath{S_1} and 
\nlbmath{S_0} are\emph{locally directly permutable} \udiff\
{replacing an occurrence of
\bigmath{S_0\over{S_l\quad S_1\quad S_r}} in a closed proof tree 
(where \math{S_1} is also applicable instead of \nlbmath{S_0}) 
with \bigmath{S_1\over{{S_0\over S_l}\quad S_0\quad {S_0\over S_r}}} 
results---{\em\mutatismutandis}---in a closed proof tree.}
\yestop\item
From the viewpoint of philosophy of mathematics it is bad practice to 
become too concrete with intuitively clear notions. 
For example, we should not say precisely which set theory we use on the 
meta-level
as long as \ZFlong, \NBGlong, \NFlong, \MLlong, \tarskigrothendieck\ and
non-wellfounded set theories \cite{aczel,vicious-circles}
\etc\ all \nolinebreak satisfy our \nolinebreak needs.
Although the case of permutability is not as self-evident as the case of 
set theory, the low 
rigor of our notion of permutability was sufficient until now.
Indeed, there is no definition of permutability or 
\nonpermutability\ in \wallen's whole book \cite{wallen},
although the avoidance of \nonpermutability\ is one of its main 
subjects, \cf\ \sectref{section introduction}.
\yestop\item
My formalization of the notion of permutability depends on the 
notions of a\emph{principal meta-variable} of an\emph{inference rule} and
is somewhat technical and difficult, 
even in the rudimental form we will present
below.
\end{enumerate}

\yestop\noindent
To avoid clutter, we define permutability only for sequent calculi.
The definition for tableau calculi is analogous.
Formally, for each inference rule, we have to define which meta-variables
are principal and which are not. 
On the one hand, the meta-variables of the principal formulas 
have to be principal, and an instantiation of all principal meta-variables
must determine the existence of an instantiation of the other meta-variables
such that the inference rule becomes applicable.
On the other hand, it is not appropriate to define all meta-variables of 
an inference rule to be principal, because this results in a 
general \nonpermutability\ of inference steps.

\begin{definition}[Principal Meta-Variables]\label
{definition meta-variables}\mbox{}
\\In our inference rules of \figuref{figure reductive rules} in
\sectref{section inference rules} exactly
the meta-variables \math A, \math B, \math x, \math t, 
\wforallvari x{}, and \sforallvari x{}
are\emph{principal}; and the other meta-variables, 
\ie\ \math\Gamma, \math\Pi, are not principal.
In lemma application steps as explained in \sectref{section lemma application},
the \nlbmath{A_k} and \nlbmath{C_i} are principal, whereas the \nlbmath{B_j}
are not.
For technical simplicity, we ignore our definitional expansion steps
on \nlbmath\forall, \nlbmath\exists, \nlbmath\lim, assuming a complete
expansion at the calculus level.
\yestop\yestop\end{definition}

\begin{definition}[Inference Step]\label
{definition inference step}\mbox{}
\par\noindent A\emph{proof tree} 
is a labeled tree whose root is labeled with a sequent and
whose paths are labeled with sequents and inference steps alternately, 
such that there is a proof history of applicable inference steps
(expansion steps) 
and global 
applications of \math R-substitutions on \fev s
(which instantiate the \fev s of their domains in all
occurrences in all
labels of the proof tree, \ie\ in all sequents\emph{and in all
inference steps}),
starting from a proof tree consisting only of a root node.
(Of course, the parent and child nodes of a node labeled with an inference
 step must be labeled with the conclusion and the premises 
 of this inference step, respectively.)
\par\noindent\label{definition closed proof tree}%
A proof tree is\emph{closed} \udiff\ 
all its leaves that are not labeled with inference steps 
are labeled with axioms.
\par\noindent An\emph{inference step} is a triple \trip I\pi\varrho\
labeling a node in a proof tree
where \math I is an inference rule and \math\pi\ and \math\varrho\
are substitutions of the principal and non-principal meta-variables 
of \nlbmath I, respectively; \ 
so that \bigmaths{I\inpit{\pi\tight\uplus\varrho}}{} describes the inference
step with parent (conclusion) and child (premise) 
nodes as an instance of the inference rule \nlbmath I.\end{definition}

\halftop\yestop\yestop\noindent
Note that in \defiref{definition inference step} we indeed have to 
refer to the proof history because the 
\mbox{\deltaplus-step} %s \app{\delta_0^+}{\sforallvari\delta f} and
\app{\delta_0^+}{\sforallvari\delta g} applied to 
\nlbmath{(1^5.3.1)} at the beginning of \sectref{section backtracking}
would not be admitted if \nolinebreak we applied the \math R-substitution 
\nlbmath\sigma\ before expanding the proof tree by the \deltaplus-step. 
This \nolinebreak is 
because \mbox{\deltaplus-steps} have to introduce\emph{new} \fuv s, and
\nlbmath\sigma\ would already introduce %\sforallvari\delta f \nolinebreak and 
\nolinebreak\sforallvari\delta g before.

\halftop\yestop\yestop\noindent
Roughly speaking, permutability of two steps \nlbmath{S_1} and 
\nlbmath{S_0} simply means the following:
{\em In a closed proof tree where 
 \nlbmath{S_0} precedes \nlbmath{S_1}
 and where \nlbmath{S_1} was already applicable before \nlbmath{S_0},
 we can do the step \nlbmath{S_1} before \nlbmath{S_0}
 and find a closed proof tree nevertheless.}

\halftop\yestop\yestop\begin{definition}[Permutability]\label
{definition permutability}\mbox{}
\par\yestop\noindent 
Let \trip{I_1}{\pi_1}{\varrho_1} and 
\trip{I_0}{\pi_0}{\varrho_0} be two inference steps.
\par\yestop\noindent 
\trip{I_1}{\pi_1}{\varrho_1} and 
\trip{I_0}{\pi_0}{\varrho_0} are\emph
{permutable for a given threshold \nlbmath m for \mbox
{\math\gamma-multiplicity}}
\ \udiff\ 
\par\noindent for any closed 
proof tree \nlbmath{T\,} with \math\gamma-multiplicity \nlbmath m
satisfying that\begin{enumerate}\noitem\item
\math{n_i} is an inference node in \nlbmath T
labeled with \nlbmath{\trip{I_i}{\pi_i}{\varrho_i}}, for \math{i\in\{0,1\}}, 
\noitem\item\math{n_0,n_1} are, in this order and 
with only a sequent node in between, on the same path in
\nlbmath T from the root to a leaf, and
\noitem\item\sloppy there is a substitution \math{\phi} such that the 
parent sequents (conclusions) of
\nlbmath{{I_0}\inpit{\pi_0\tightuplus\varrho_0}} 
and of
\nlbmath{{I_1}\inpit{\pi_1\tightuplus\phi}} 
are identical;\noitem\end{enumerate}
there is a closed proof tree with \math\gamma-multiplicity \nlbmath m
which differs from \nlbmath T only in the subtree starting with \nlbmath{n_0}
and the root label of this subtree is \nlbmath{\trip{I_1}{\pi_1}{\phi}}.
\par\yestop\noindent 
\trip{I_1}{\pi_1}{\varrho_1} and 
\trip{I_0}{\pi_0}{\varrho_0} are\emph{permutable} \udiff\ they are
permutable for any given threshold \math{m\in\N} of \math\gamma-multiplicity.
\par\yestop\noindent 
\math{I_1} and \nlbmath{I_0} are\emph{generally permutable} \udiff\ 
all inference steps of the forms
\trip{I_1}{\pi_1}{\varrho_1} and 
\trip{I_0}{\pi_0}{\varrho_0} are 
permutable.\end{definition}\vfill\pagebreak

\figurecompleteproof h\pagebreak

\begin{example}\label{example permutability}\\
For inferring the \nonpermutability\ of \math\beta\ and \deltaplus\ from
\lemmref{lemma no completion}, we have to instantiate \defiref
{definition permutability} as follows:
\par\yestop\noindent\mbox{}\hfill\math{
\begin{array}{@{}l@{}c@{}l@{}}
  n_0
 &\approx
 &(1^5.3.1)\red(1^5.3.1^2)\mbox{\ \ (\cf\ \sectref{section backtracking})}
\\I_0
 &\,\,\,\mbox{is}\,\,\,
 &\pair\deltaplus{\neg\exists}\mbox
  { of \figuref{figure reductive rules} in \sectref{section inference rules}}
\\\pi_0
 &=
 &\left\{\begin{array}[c]{@{}l l l}
    \boundvari   x{}
   &\mapsto
   &\boundvari  \delta g;
  \\\sforallvari x{}
   &\mapsto
   &\sforallvari\delta g;\footroom
  \\A
   &\mapsto
   &\inparentheses{
    0\tight<\boundvari\delta g
    \und
    \exists\boundvari x g\boldunequal\wforallvari x 0\stopq\inparenthesesoplist{
        |\app{\wforallvari g{}}{\boundvari x g}
         \tight-\wforallvari y g|<{\wforallvari\varepsilon{}\over 2}
      \oplistantiimplies
        |\boundvari x g\tight-\wforallvari x 0|<\boundvari\delta g
    }}
  \\\end{array}\right\}
\\\varrho_0
 &=
 &\left\{\begin{array}[c]{@{}l@{\,\,}l@{\,\,}l}
    \Gamma
   &\mapsto
   &\inparenthesesoplist{
      0\tight<\min(\sforallvari\delta f,\sforallvari\delta g)
    \oplistund
    \forall x\boldunequal\wforallvari x 0\stopq 
    \inparenthesesoplist{ 
        \left|\begin{array}{@{}l@{}}
           \inpit{\app{\wforallvari f{}}{x}\tight+\app{\wforallvari g{}}{x}}
         \\-\inpit{\wforallvari y f\tight+\wforallvari y g}
	 \\\end{array}\right|
        <\wforallvari\varepsilon{}
      \oplistantiimplies
        |x\tight-\wforallvari x 0|
       <\min(\sforallvari\delta f,\sforallvari\delta g)
    }
    }\comma\ldots;
  \\\Pi
   &\mapsto
   &\ldots
  \\\end{array}\right\}
\\\end{array}}
\par\yestop\mbox{}\hfill\noindent\LINEmath{
\begin{array}{@{}l@{}c@{}l@{}}
  n_1
 &\approx
 &\begin{minipage}[t]\textwidthminustwocm
  {``a new step of an alternative closed proof tree that results from the
     closed proof tree of \sectref{section backtracking} by permuting the 
     \math\beta-step at \nlbmath{(1^5.3.1^2)} and the 
     steps \nlbmath{\alpha^2,\app{\gamma_0}{\sforallvari x{}}^2}
     applied to \nlbmath{(1^5.3.1)}. 
     This alternative proof tree is depicted in 
     \figurefsix\ above.
     (For pedagogical reasons only,
     we delayed the potentially sinful
     \math\beta-step until we were forced to do it.)''}
  \end{minipage}
\\I_1
 &\,\,\,\mbox{is}\,\,\,
 &\pair\beta{\tight\wedge}\mbox
  { of \figuref{figure reductive rules} in \sectref{section inference rules}}
\\\pi_1
 &=
 &\left\{\begin{array}[c]{@{}l l l@{}}
    A
   &\mapsto 
   &0\tight<\min(\sforallvari\delta f,\sforallvari\delta g);\footroom
  \\B
   &\mapsto 
   &    \forall x\boldunequal\wforallvari x 0\stopq 
    \inparenthesesoplist{
        \left|\begin{array}{@{}l@{}}
           \inpit{\app{\wforallvari f{}}{x}\tight+\app{\wforallvari g{}}{x}}
         \\-\inpit{\wforallvari y f\tight+\wforallvari y g}
	 \\\end{array}\right|
        <\wforallvari\varepsilon{}
      \oplistantiimplies
        |x\tight-\wforallvari x 0|
       <\min(\sforallvari\delta f,\sforallvari\delta g)
    }
  \\\end{array}\right\}
\\\end{array}
}
\yestop\halftop\end{example}

\noindent
Now, the \nonpermutability\ 
of the critical \math\beta- and \deltaplus-steps of 
\examref{example permutability}
follows from \lemmref{lemma no completion},
because there is no alternative proof tree
which differs only in the subtree starting at \math{n_0} and having a new 
subtree there starting with the critical \nlbmath\beta-step. \
The deeper reason for this is 
that the instantiated \fev s occur outside the subtree of the \deltaplus-step,
\cfnlb\ \sectref{section escape their quantifiers scopes}. \  
According to \lemmref{lemma no completion}, 
there is no proof of \nlbmath{(1^5.1)}, 
\nlbmath{(1^5.2)} and \nlbmath{(1^5.3.1.1)} with the instantiation by
\nlbmath\sigma\ given by the failed proof attempt. \
Since the partial instantiation by \nlbmath\sigma\ agrees with the
full instantiation in the closed proof tree of the successful proof
of \figurefsix,
we have the required witness for the \nonpermutability\ of 
\math\beta\nolinebreak\ and \nolinebreak\deltaplus, indeed.
Thus, as corollaries we get:

\begin{corollary}
On a threshold for \mbox{\math\gamma-multiplicity of \math{\,1}}, \ 
the inference steps\par\LINEnomath{% 
\trip{\pair\beta{\tight\wedge}}{\pi_1}{\varrho_1} \ \ and \ \ 
\trip{\pair\deltaplus{\neg\exists}}{\pi_0}{\varrho_0}}\par\noindent
(as labels of the nodes \nlbmath{n_1} and \nlbmath{n_0}, \resp)
as given in \examref{example permutability}
are not permutable.\end{corollary}
\begin{theorem} \ 
\math\beta-\ and\/ \deltaplus-steps are not generally permutable,
\begin{itemize}\noitem\item 
neither in the sequent calculus of \cite{wirthcardinal} 
(\cf\ our \figuref{figure reductive rules} 
 in \sectref{section inference rules}),
\halftop\notop\item nor in standard 
free-variable tableau calculi with \deltaplus-rules as the only 
\math\delta-rules, such as the ones in \cite{fitting,deltaplus}.
\cleardoublepage\end{itemize}\end{theorem}

\section{Conclusion}\label{section conclusion}
Even with more liberalized \math\delta-rules available today
(such as \deltaplusplus-, \deltastar-, \deltastarstar-, and 
\math{\delta^\varepsilon}-rules,
\cfnlb\ \sectref{section where all the different deltas are referenced}), 
the \deltaplus-rules stay important, 
both conceptually and for stepwise presentation 
and limitation of complexity in teaching, research, and publication. 
For instance, the \mbox{\deltaplus-rules} are the free-variable tableau 
rules used in the current edition of 
\fitting's excellent textbook \nlbcite{fitting}. \ 
Moreover, until very recently \nlbcite{delta-generic} 
nobody realized that 
the \deltastar- and \deltastarstar-rules were unsound in their 
original publications (\incl\ their corrigenda!).

When the \deltaplus-rules occurred first in \nlbcite{deltaplus},
they \nolinebreak seemed so simple and straightforward. 
Today, a dozen years later,
they \nolinebreak are still not completely understood. 
We have shown that the \deltaplus-rules have unrealized properties yet,
such as the \nonpermutability\ of \math\beta- and \deltaplus-steps.
Indeed, there are several\emph{open problems}, 
such as, from theoretical to practical:

\subsection{Complexity?}
Does the non-elementary reduction in proof size \nlbcite{baazdelta}
from the \deltaminus- to the \deltaplusplus-rules 
mean a non-elementary reduction in proof size from \deltaminus\ to \deltaplus,
or from \deltaplus\ to \deltaplusplus\ 
(exponential at least \nlbcite{deltaplusplus}),
or both?

\subsection{More \NonPermutabilities?}%
Why was the \nonpermutability\ of \math\beta\ and \nolinebreak\deltaplus\ 
not noticed before? 
May there be others around? 

\subsection{Optimization?}Although the \nonpermutability\ of 
\math\beta- and \mbox{\deltaplus-steps} is not visible with
non-liberalized \mbox{\math\delta-rules}
and not serious in theory with further liberalized \mbox{\math\delta-rules},
it \nolinebreak is always present and of major importance in practice;
both for efficiency of proof search and for human-oriented proof presentation.
The same holds for
the optimization problem of finding a good order of application
for the \math\beta-steps. \ 

\subsection{Are the known notions of Completeness relevant in practice?}
The mere existence of a proof is not sufficient for \maslong s, 
where we need the existence of a
proof that closely mirrors the proof the mathematician interacting with the 
system has in mind, searches for, or \nolinebreak plans.

Freshmen who think that the \deltaminus-rules would admit 
human-oriented proof construction should try to do the 
proof of {\nolinebreak$($$\lim$$+$$)$} 
with the \deltaminus-rules as the only \math\delta-rules.
There will be more reasons and occasions
to use the presentation of this complete and interesting example proof 
for further reference!

I \nolinebreak must admit, however, 
that I do not know how to grasp a practically relevant 
notion of completeness. \ 
The sequent calculus of our inductive theorem prover \QUODLIBET\ 
\nlbcite{quodlibet-cade} has been improved over a dozen years 
of practical application to admit our proofs; 
and still needs and gets further improvement.

 The automatic generation of a non-trivial proof
 for a given input conjecture
 is typically not possible today
 and probably will never be. Thus, besides some rare exceptions---as the 
 automation of proof search 
 will always fail on the lowest logic level from time to time---the only 
 chance for automatic theorem proving to become
 useful for mathematicians is a synergetic interplay 
 between the mathematician and the machine.
 For this interplay---to give the human user a chance to interact---the 
 calculus\emph{itself} must be human-oriented.
 Indeed, it does not suffice to compute human-oriented representations;
 not in the end, and---as the syntactical problems have to be 
 presented accurately---also not intermediately in a user interface.

Thus, also the possibility to overcome the 
\nonpermutability\ of \math\beta\nolinebreak\ and \nolinebreak\deltaplus\ 
by replacing the \deltaplus-rules 
with \deltaplusplus-rules 
as described in \sectref{section deltaplusplus} is not adequate 
for human-oriented reasoning,
for which we need 
matrix calculi and indexed formula trees \nlbcite{sergediss,wallen}
to admit a lazy sequencing of \math\beta-steps,
so that the connection-driven path construction may tell us in the end,
which sequencing of the \math\beta-steps we need.\footnote
{An anonymous 
 referee of a previous version of this \daspaper\ wrote:
 \begin{quote}
 ``The arguments against the use of \deltaplusplus\ 
  (that the proofs found this 
  way are not human-oriented) are not convincing. 
  It is well-known that improved
  Skolemization rules can be simulated with applications of the cut rule. 
  So one could proceed as follows. 
  Use \deltaplusplus\ for proof generation, for 
  presentation insert the respective cut steps. This way any forms of 
  sophisticated Skolemization could be replaced by case distinctions, 
  which are 
  easily understandable by any human user.'' \end{quote}
 The point that is missed in this critique is the following.
 The automatic generation of non-trivial proofs is typically not possible today
 and probably will never be. Thus, besides some rare exceptions---as the 
 automation of proof search 
 will always fail on the lowest logic level from time to time---the only 
 chance for automatic theorem proving to become
 useful for mathematicians is a synergetic interplay 
 between the mathematician and the machine.
 For this interplay---to give the human user a chance to interact---the 
 calculus\emph{itself} must be human-oriented.
 Thus, it does not suffice to compute human-oriented representations;
 not in the end, and---as the syntactical problems have to be 
 presented accurately---also not intermediately in a user interface.}

\subsection{Is Soundness sufficient in practice?}
The notion of\emph{safeness} (soundness of the
reverse inference step, for failure detection after generalization,
\eg\nolinebreak\ for induction) seems to become standard 
\nlbcite{sergetableau,isabellehol,wirthdiss,wirthcardinal}. \
And in \cite{wirthgreen,wirthcardinal} we have also added 
the notion of\emph{preservation of solutions}. 
This means that 
the closing substitutions on the rigid variables of the sub-goals
must solve the input theorem's
rigid variables, which make sense as placeholders
for concrete bounds and side conditions 
of the theorem which only a proof can tell. \ 

\subsection{Conclusion}
Although more useful for proof search in classical logic than \hilbert\ 
\nlbcite{grundlagen} and
\ND\ calculi \nlbcite{gentzen},
sequent \nlbcite{gentzen} and tableau calculi \nlbcite{fitting}
are still not adequate for 
a synergetic interplay of 
human proof guidance and automatic proof search 
\nlbcite{wirthcardinal},
which we hope to achieve with matrix calculi such as 
\CORE\ \nlbcite{sergediss}. \

As \nolinebreak the automation of proof search 
will always fail on the lowest logic level from time to time,
be aware:
{\em The fine structure and human-orientedness of a calculus
does matter in practice!}

\vfill\pagebreak
\yestop\yestop\yestop\yestop
\mysection{Acknowledgements}

\yestop\noindent
I have to thank 
the anonymous referees of previous versions of this \daspaper\
for the useful elements of their critiques. \ 

\yestop\noindent
I would like to thank 
my co-lecturers for giving the sometimes better 
and always less exhausting lectures in our course \nlbcite{maslecture},
and the students of the course \nlbcite{maslecture} and especially 
its predecessor \nlbcite{wirthlecture} 
(who were the first to suffer from my formalization of the 
(\math{\lim\tight+}) \nolinebreak proof)
for teaching each other and sharing all those joys of logic. \ 
This means that I would like to thank---among others---\autexiername, 
\benzmuellername, Mark Buckley, \dietrichname,
\fiedlername, \huttername, 
\meiername, \polletname, Marvin Schiller, 
\samoaname, \siekmannname, 
\stephansecondnamename, Fabian M. Suchanek, \wagnername, and \wolskaname.

\yestop\noindent
Last but not least,
I do thank \brownchadname\ very much indeed for giving me the
most careful and constructive comments and suggestions for improvement
I ever got in my life. \ 
% \begin{verbatim}
% \begin{slight irony}
% \end{verbatim}
% Since my wonderful mathematics teacher 
% \bruggaiername\ 
% at high-school a quarter of a century ago, 
% nobody taught me mathematical rigor so wonderfully as Chad did:
% ``Maybe the mistake was the order of the rules after the bad beta-step?
% It wasn't, but nevertheless.''
% \begin{verbatim}
% \end{slight irony}
% \end{verbatim}
A comparison of his report with one of the anonymous ones of 
\thefourteenthTABLEAUfive, (of a previous version of this \daspaper)
suggests that 
new forms of evaluation that further science
by communication between scientists are in great demand. 
I would like to dedicate this paper to Chad,
for various reasons.
% With the current standard, 
% science managers
% who do not want to read something new which may
% endanger their own scientific school can easily abuse their anonymity and 
% hinder scientific progress.

\knuthquotation{\fraknomath Hie ist Wei\esi heit. Wer verstand hat/der uberlege}
{\cite[Offenbarung\,XIII]{bibliagermanica}}
%\vfill\pagebreak

\yestop\yestop\yestop\yestop\mysection{Notes}\halftop\halftop\yestop\yestop
\begingroup
\theendnotes
\endgroup

\vfill\pagebreak

\end{document}